\pdfoutput=1

\documentclass[11pt]{article}

\usepackage{acl}

\usepackage[T1]{fontenc}
\usepackage{pifont}

\usepackage{times}
\usepackage{soul}
\usepackage{url}
\usepackage[utf8]{inputenc}
\usepackage{graphicx}
\graphicspath{ {.} }
\usepackage{amsmath}
\usepackage{amsthm}
\usepackage{booktabs}
\usepackage{algorithm}
\usepackage{algorithmic}
\usepackage{amsmath}
\usepackage{bbm}
\usepackage{amsfonts}
\usepackage{multirow} %% Pour mettre un texte sur plusieurs rangées
\usepackage{multicol} %% Pour mettre un texte sur plusieurs colonnes
\usepackage{calc}
\usepackage[colorinlistoftodos]{todonotes}
\usepackage{tabularx}
\usepackage{xcolor}
\usepackage{xspace}
\usepackage{microtype}
\usepackage{inconsolata}
\usepackage{latexsym}

\newcommand{\fydone}[1]{}

\definecolor{LightGray}{rgb}{0.9,0.9,0.9}

\DeclareMathOperator{\ath}{arctanh}

\newcommand{\simi}{\mathrm{sim}}
\newcommand{\src}{\ensuremath{\mathbf{x}}}
\newcommand{\tgt}{\ensuremath{\mathbf{y}}}

\newcommand{\Lev}{\ensuremath{\operatorname{Lev}}}
\newcommand{\LED}{{LED}}

\newcommand{\BLEU}{\texttt{BLEU}\xspace}
\newcommand{\COMET}{\texttt{COMET}\xspace}

\newcommand{\mlevt}[1]{\texttt{TM$^{#1}$-LevT}}
\newcommand{\NFA}{\texttt{NFA}\xspace}
\newcommand{\EuroLLM}{\texttt{EuroLLM}\xspace}
\newcommand{\rtvr}[1]{\textbf{\texttt{#1}}}
\newcommand{\base}{\rtvr{dense}\xspace}
\newcommand{\basebow}{\rtvr{dense+bow}}
\newcommand{\fuzzysrc}{\rtvr{fuzzy-src}\xspace}
\newcommand{\fuzzybt}{\rtvr{fuzzy-bt}\xspace}
\newcommand{\fuzzygold}{\rtvr{fuzzy-gold}\xspace}
\newcommand{\LASER}{\rtvr{LASER}\xspace}
\newcommand{\LaBSE}{\rtvr{LaBSE}}
\newcommand{\ft}{\rtvr{ft-}}
\newcommand{\MSE}{\rtvr{-MSE}\xspace}
\newcommand{\MAE}{\rtvr{-MAE}\xspace}
\newcommand{\Rank}{\rtvr{-Rank}\xspace}
\newcommand{\Spval}{$^*$}

\newcommand{\sm}[1]{\small#1\normalsize}
\newcommand{\tn}[1]{\scriptsize#1\normalsize}
\newcommand{\contamnllb}[1]{\underline{#1}}

\title{Improving Retrieval-Augmented Neural Machine Translation with Monolingual Data} %

\author{
	Maxime Bouthors$^{\dagger}$ \quad Josep Crego$^{\dagger}$ \quad Dakun Zhang$^{\dagger}$ \quad François Yvon$^{\ddagger}$ \\
	$^{\dagger}$SYSTRAN by ChapsVision, 5 rue Feydeau, F-75002 Paris, France \\
	$^{\ddagger}$Sorbonne Université, CNRS, ISIR, F-75005 Paris, France \\
	\texttt{\{mbouthors, jcrego, dzhang\}@chapsvision.com} \\
	\texttt{francois.yvon@cnrs.fr}
}

\begin{document}
% TODO:
% - [x] word-level matching
% - [x] distances are monolingual
% - [x] related work
% - [ ] example with various retrievers in the appendix, including terms, domain effect, this can be discussed in text
% - [ ] Distillation and LevT
% - [X] Other approaches to BT (in related work Immamura)
% - [X] limitation section: no low resource, high resource use cases
\maketitle

\begin{abstract}
	Conventional retrieval-augmented neural machine translation (RANMT) systems leverage bilingual corpora, e.g., translation memories (TMs). Yet, in many settings, monolingual corpora in the target language are often available. This work explores ways to take advantage of such resources by directly retrieving relevant target language segments, based on a source-side query. For this, we design improved cross-lingual retrieval systems, trained with both sentence level and word-level matching objectives. In our experiments with three RANMT architectures, we assess such cross-lingual objectives in a controlled setting, reaching performances that match those of standard TM-based models. We also showcase our method on a real-world settings, using much larger monolingual corpora,
	and observe strong improvements over both the baseline setting and general-purpose cross-lingual retrievers.
\end{abstract}

\section{Introduction \label{sec:introduction}}

% Example \shortcite{bouthors-etal-2023-towards}; \citep{bouthors-etal-2023-towards}; \cites{bouthors-etal-2023-towards},

The use of Retrieval-Augmented Generative Models \citep{li-etal-2023-survey} is rapidly expanding, owing to their built-in ability to condition generations with illustrative segments retrieved from memory. The combination of (machine) retrieval and (human) regeneration has long been a key principle of Computer Aided Translation (CAT) systems \citep{arthern-1978-machine,kay-1997-proper,bowker-2002-computer}:  segments resembling the source sentence are first retrieved from Translation Memories (TM), providing translators valuable suggestions that can then be edited into a new translation. Such techniques have been transposed in Statistical MT \citep{koehn-senellart-2010-convergence}, more recently in Retrieval-Augmented Neural Machine Translation (RANMT) e.g., \citep{gu-etal-2018-search}, making the entire process fully automatic. Most subsequent work has continued to leverage parallel data\fydone{cut: in the form of TMs}, performing retrieval in the source language, then using the retrieved target side text in generation. Similar principles underlie few-shot MT with large language models, where both the source and target sides of examples are inserted into the prompt \citep{brown-etal-2020-language}.

% Framing is a bit different from the abstract, where we insist on using alternative sources of data
% Also: insist on lexical match
TM-based approaches typically retrieve examples with \textbf{lexical fuzzy matchers}, using BM25 and/or the Levenshtein distance (\LED) as filters or rankers \citep{bouthors-etal-2024-retrieving}. These are known to be computationally efficient and to often surpass continuous-space retrievers \citep{xu-etal-2020-boosting}: this is because lexical (source) matches often come with appropriate translation of terms. However, using source-side lexical similarity raises two issues that may cause inadequate segments to be retrieved: (a) noisy alignments between source and target side of a parallel instance; (b) intrinsic linguistic divergences between the two languages \citep{dorr-1994-machine}.
% Complement
% \option{Lexical matches also fail to retrieve semantically relevant paraphases that would better be captured by neural retrievers.}

These problems can be sidestepped by performing \textbf{retrieval directly on the target side}, using Cross-Lingual Information Retrieval (CLIR) techniques \citep{cai-etal-2021-neural,tamura-etal-2023-target}. An important additional benefit is to dispense with the use of parallel data and potentially inform the generation process with relevant monolingual resources. As demonstrated by the recurrent use of target side monolingual data thanks to back-translation \citep{sennrich-etal-2016-improving}, such resources are much easier to find than parallel ones in many domains, and are also devoid of so-called \emph{translationese phenomena} \citep{bogoychev-sennrich-2020-domaintranslationese}. As such, monolingual corpora constitute a very useful, yet under-studied, source of high-quality segments for RANMT. To search such resources, cross-lingual retrievers are readily available thanks to multilingual sentence encoders such as LASER \citep{artetxe-schwenk-2019-massively} or LaBSE \citep{feng-etal-2022-language}. As they only rely on dense representations, such retrievers however often fail to retrieve lexically relevant examples, i.e, examples that would contain translations of the exact source words.\fydone{convoluted, but there can not be any overlap with the source as there is no source !}
% \citep{zebaze-etal-2024-context}.
% There already exists language agnostic sentence encoders that can be used in a \emph{source-to-target} setup \citep{zebaze-etal-2024-context}. It appears that this setup is very little studied.

In this work, we propose novel methods to boost the lexical matching abilities of language-agnostic sentence encoders, thereby improving their effectiveness for RANMT. For this, we consider various ways to fine-tune a pre-trained encoder into reproducing the behaviour of lexical fuzzy matchers.
% exploring the use of various dedicated training losses.
We experiment with three language pairs (English-French, German-English and English-Ukrainian) and multiple domains of various sizes and diversities.\fydone{or three} We consider three types of RANMT systems: an autoregressive source-augmented transformer \citep{bulte-tezcan-2019-neural}, a non-autoregressive edit-based model \citep{bouthors-etal-2023-towards} and a Large Language Model (LLM), with in-context ($k$-shot) learning \citep{brown-etal-2020-language}.
We first study the performance of CLIR techniques in a controlled setting, showing that they match standard TM-based models with source-side retrieval.
% \textbf{\tomodif{vastly surpass standard TM-based models with source-side retrieval}}.
We then consider much larger setups, where the target monolingual resources far exceed the amount of parallel data and observe large improvements using our new techniques, which outperform various baseline systems, as well as RANMT architectures relying on generic cross-lingual retrievers.

% We train LaBSE \citep{feng-etal-2022-language} to make it predict an estimation of the Levenshtein similarity between an example and a latent estimation of the reference.
% \option{It requires a TM at training time, but we show that it can seamlessly scale to an in-domain large monolingual corpus at inference time, with high performance gains. --> Limitations}
% \option{Use of fine-tuned LLMs -> Limitations; check domain effect of the lexical matching ; Need Parallel data in training}

\section{Related Work \label{sec:related}}

% CAT
\paragraph{Retrieval Augmented NMT}
Our work falls under the scope of Retrieval-Augmented Generation, leveraging both lexical (BM25 \citep{robertson-walker-1994-some}, Levenshtein distance \citep{levenshtein-1965-binary}) and semantic methods for the Information Retrieval (IR) component. Semantic methods usually rely on dual encoders trained on parallel data \citep{gillick-etal-2018-end} with a contrastive loss \citep{sohn-2016-improved}. Our focus is on massively multilingual sentence encoders that encode mutual translations as close neighbors, such as LASER \citep{artetxe-schwenk-2019-massively} and LaBSE \citep{feng-etal-2022-language}.

In recent years, retrieval-augmented MT systems have received increasing attention. One key motivation is to enhance model transparency by providing users with the retrieved examples that inform the generated output \citep{rudin-cynthia-2019-stop}. A simple implementation of this idea encodes the target side of the example(s) along with the source sentence, providing an additional translation context \citep{gu-etal-2018-search,bulte-tezcan-2019-neural,xia-etal-2019-graph,he-etal-2021-fast,cheng-etal-2022-neural,agrawal-etal-2023-context}. It is also possible to leverage both the source and target sides of the retrieved example(s) \citep{pham-etal-2020-priming,reheman-etal-2023-prompting}.

Rather than regenerating a complete translation with extended context, edit-based models perform minimal changes to the retrieved examples and generate their output \emph{in a non-autoregressive (NAR) fashion}. Such models are studied in \citep{niwa-etal-2022-nearest,xu-etal-2023-integrating,zheng-etal-2023-towards}, adapting the Levenshtein Transformer \citep{gu-etal-2019-levenshtein}. \citet{bouthors-etal-2023-towards} generalize this approach to simultaneously handle multiple examples.

MT systems based on Large Language Models (LLMs) can also seamlessly accomodate examples through In-Context Learning (ICL), where the LLM prompt is enriched with \emph{demonstrations of the translation task}, comprising both the source and target sides of parallel samples \citep{radford-etal-2019-language}. Several studies have tried to optimize the performance of such architectures for MT, examining the impact of prompt modifications, the number of demonstrations and the retrieval procedure \citep{moslem-etal-2023-adaptive,vilar-etal-2023-prompting,zhang-et-al-2023-prompting,hendy-etal-2023-howgood,bawden-yvon-2023-investigating,zebaze-etal-2024-context}.

\paragraph{Monolingual Data in NMT} Augmenting NMT models with target side monolingual data by back-translating segments is a well-established strategy for improving model training \citep{sennrich-etal-2016-improving, edunov-etal-2018-understanding}, already considered in RANMT setup \citep{tezcan-etal-2024-improving}.\fydone{Also add in final version: Minh} \citep{cai-etal-2021-neural,tamura-etal-2023-target} dispense with back-translation, performing directly the retrieval in the target language with CLIR techniques: the latter uses existing retrievers, while the former jointly trains the retriever and the MT components.

\paragraph{Metric Learning} Our lexical alignment objectives (\textsection~\ref{ssec:training}) corresponds to a form of neural metric learning, where a high-dimensional metric space is reduced to a smaller one \citep{suarez-etal-2021-atutorial}. This line of work aims to learn distances from the data \citep{kulis-2013-metric,bellet-etal-2015-metric}, which is particulary useful for similarity-based algorithms used in IR \citep{cakir-etal-2019-deep}.

\section{Method \label{sec:method}}

\subsection{Problem Formulation \label{ssec:formulation}}

Given a source sentence $\src$ and a corpus of target segments $\mathcal D = (\tgt_1, \dots, \tgt_N)$, the goal of CLIR is to retrieve $k$ \emph{relevant} examples: $(\tilde \tgt_1, \dots, \tilde \tgt_k)$.

Our approach closely follows neural-based cross-lingual retrievers, and comprises two main steps. First, $\mathcal D$ is encoded with a multilingual sentence encoder $E_\theta$. Then, for any query source sentence $\src$, retrieval is performed as a $k$NN similarity search in the embedding space, using the cosine similarity:
\begin{equation}
	\simi_\theta (\src, \tilde \tgt) = \cos(E_\theta(\src), E_\theta(\tilde \tgt)).
\end{equation}

The quality of the retrieved sentences only depends on the choice of $E_\theta$. Usually, $E_\theta$ is trained as a Siamese encoder with a contrastive loss on parallel data, a usual IR objective \citep{sohn-2016-improved} that primarily captures semantic similarity.
However, it is desirable for the retrieved $\tilde{\tgt}_i$ instances to exhibit lexical similarities with the unknown translation $\tgt$ of the source $\src$. In specialized domains (like law or medicine), adherence to domain-specific terminology takes precedence over semantic similarity, thereby motivating the design of our proposed lexical-based alignment objective. To address this need, we study various fine-tuning objectives based on the \emph{Levenshtein similarity}.
These objectives are specifically designed \textbf{to boost the retrieval of segments with a large numbers of lexical matches in the source query}, by learning a new embedding space, achieved through the optimization of parameters $\theta$, such that:\footnote{As edit distances are only computed monolingually, our method does not assume any lexical overlap between $\src$ and $\tilde \tgt$.}
\fydone{should make the transformation explicit}
\begin{equation}
	f(\simi_\theta (\src, \tilde \tgt)) \approx \Lev(\tgt, \tilde \tgt),
	\label{eq:cos-lev}
\end{equation}
where $f$ is a (potentially learnt) mapping from the codomain of the cosine function into the codomain of Levenshtein similarity function \Lev{}, defined as:
\begin{equation}
	\Lev(\tgt, \tilde \tgt) = 1 - \frac{\Delta (\tgt, \tilde \tgt)}{\max (|\tgt|, |\tilde \tgt|)}.
	\label{eq:lev-sim}
\end{equation}
$\Delta$ is the Levenshtein distance \citep{levenshtein-1965-binary}.

\subsection{Training \label{ssec:training}}
We consider three objective functions. The first two learn to predict the Levenshtein similarity \eqref{eq:lev-sim}. We choose $f$ to be a learnable function that maps $[-1, 1]$ to $[0, 1]$:
\begin{equation}
	f(t) = \sigma(a\times \ath(t) + b),
	\label{eq:mapping-function-cos-to-lev}
\end{equation}
with $a$ (slope) and $b$ (position) parameters and $\sigma$ the sigmoid function.
The loss is defined as:
\begin{equation}
	\hspace{-0.5 em} \mathcal L(\src, \tgt, \tilde \tgt) = \operatorname{Err}(f(\simi_\theta (\src, \tilde \tgt)) , \Lev(\tgt, \tilde \tgt)),
	\label{eq:loss-metric-learning}
\end{equation}
where the error function $\operatorname{Err}$ is either the mean square (MSE) or the mean absolute error (MAE):% defined as:
\begin{align}
	\label{eq:1}
	\operatorname{MSE}(x,y) = (x - y )^2 \\
	\operatorname{MAE}(x,y) = |x - y |.
\end{align}

The third objective is a ``learning to rank" objective \citep{cao-etal-2016-learning}. Given a set of $k$ retrieved segments $\tilde \tgt_{[1:k]}$ sorted in decreasing order w.r.t. $\Lev(\tgt, \cdot)$,
it computes:
\begin{align}
	\mathcal L( & \src,\tgt, \tilde \tgt_{[1:k]})  =  \sum_{i > j} \max (0, U_{ij})                     \\ \nonumber
	U_{ij}      & = \simi_\theta(\src, \tilde \tgt_i)  - \simi_\theta(\src, \tilde \tgt_j)              \\
	            & + m \times |\Lev(\tgt, \tilde\tgt_j) - \Lev(\tgt, \tilde\tgt_j)|) \label{eq:rankloss}
\end{align}
with $m$ a scaling factor. Equation~\eqref{eq:rankloss} defines a pairwise margin loss,\footnote{Inspired by the triplet loss \citep{schultz-etal-2003-advances}.} with an adaptive margin proportional to the absolute difference between the two individual \Lev{} similarities.

\section{Data and Metrics \label{sec:datametrics}}

\subsection{Data \label{ssec:datasets}}

We consider three translation directions: English into French (en-fr), German into English (de-en) and English into Ukrainian (en-uk).\fydone{or more} For our controlled experiments, we consider a wide variety of textual domains (16), using public parallel datasets from the Opus corpus \citep{tiedemann-2012-parallel}. Details are in appendix~\ref{appendix:data}.

For the large scale experiments, we expand the English-French setting with a large corpus of Wikipedia pages in French.\footnote{\url{https://huggingface.co/datasets/Plim/fr\_wikipedia\_processed}.} We split the text in pseudo-sentences via regular expression matching and remove duplicates as well as fuzzy matches from the valid/test sets\footnote{Each best match with $\Lev>0.9$ is simply removed.} of the parallel Wikipedia corpus, ending up with about $45$M French segments. The corresponding English-Ukrainian experiments, reported in Appendix~\ref{sec:en2uk}, reproduce the setting of \citet{tezcan-etal-2024-improving}, using texts in the legal domain.

\subsection{Metrics\label{ssec:metrics}}
% xSIM++ https://aclanthology.org/2023.acl-short.10.pdf
The success of cross-lingual retrieval is assessed with the average target side Levenshtein similarity (eq.~\eqref{eq:lev-sim}) between the 1-best match and the reference, for sentences in the validation set, with retrieval performed from the entire train set. We also report \texttt{xsim} error rates, which measure the abililty to identify matched sentences in a bilingual parallel corpus \citep{artetxe-schwenk-2019-margin}.

We assess translation quality with \BLEU{} \citep{papineni-etal-2002-bleu} computed with SacreBLEU \citep{post-2018-call},\footnote{signature: \texttt{nrefs:1|case:mixed|eff:no|tok:13a| smooth:exp|version:2.1.0};} as well as \COMET-20\footnote{\url{Unbabel/wmt20-comet-da}, using the defaults settings.} \citep{rei-etal-2020-comet}.
% COMET and BLEU measure different aspects of translation quality: the former seems to capture the general fluency and semantic adequacy, and is known to better correlate with human judgments \citep{rei-etal-2022-comet} than BLEU; however, it does not necessarily capture translation quality in technical domains, where the lexical matching metrics such as BLEU can also measure the appropriateness of terminological choices.

\section{Experimental Settings \label{sec:experiments}}
% \subsection{Controlled experiments}
\subsection{Retrieval Settings \label{ssec:retrieval-config}}

Our experiments consider the following retrieval methods, with up to $k=3$ closest examples:
\begin{itemize}
	\item \fuzzysrc: \emph{source-to-source} fuzzy matching ranked by their $\Lev$ similarity, as in baseline RANMT settings;
	\item \fuzzygold: \emph{target-to-target} fuzzy matching, using the same procedure as \fuzzysrc. This retriever requires access to reference target translations and \emph{corresponds to an oracle setting}, giving an empirical upper-bound of what an ideal system could achieve. It thus is only reported for the controlled experiments of \textsection~\ref{ssec:clir_results} and \textsection~\ref{ssec:trans_results};
	\item \fuzzybt: the target side of training segments are back-translated using NLLB \citep{costa-etal-2024-scaling}\footnote{We used \texttt{NLLB-200-distilled-1.3B} out-of-the-box.\\\url{https://huggingface.co/facebook/nllb-200-1.3B}} into the source language \cite{sennrich-etal-2016-improving}, then used for retrieval as in \fuzzysrc. This simulates an alternative to cross-lingual retrieval in situations where only target side examples are available;\footnote{Back-translation is further discussed in \textsection~\ref{sec:backtrans}.}
	\item \base: a BERT base \cite{devlin-etal-2019-bert} model for \emph{source-to-target} retrieval. It is trained from scratch with a contrastive loss \citep{sohn-2016-improved}. The loss $\mathcal{L} (\mathcal B)$ for batch $\mathcal B=((\src_1, \tgt_1),\dots,(\src_n, \tgt_n))$ is defined as:
	      \begin{equation}
		      - \sum_i \log \frac{\exp(\simi(\src_i, \tgt_i))}{\sum_j \exp(\simi(\src_i, \tgt_j))};
		      \label{eq:contrastive-loss}
	      \end{equation}
	\item \basebow: we add a bag-of-word loss to \base as in \citep{cai-etal-2021-neural} to enforce lexical information in cross-lingual encodings. Given a pair of parallel sentences $(\src, \tgt)$, the loss $\mathcal L (\src, \tgt)$ is computed as:
	      \begin{equation}
		      %   \mathcal L (\src, \tgt) =
		      -\sum_{w \in \mathcal Y} \log p_{\theta_1} (w|\src) - \sum_{w \in \mathcal X} \log p_{\theta_2} (w|\tgt),
		      \label{eq:bow-loss}
	      \end{equation}
	      where $\mathcal X$, $\mathcal Y$ are the bag-of-words for $\src$ et $\tgt$; $p_{\theta_1}$ and $p_{\theta_2}$ are distinct linear projections of $E_\theta(\src)$ and $E_\theta(\tgt)$;
	\item \LASER: we use LASER \citep{artetxe-schwenk-2019-massively} as a first pre-trained \emph{source-to-target} retriever. LASER derives from a multilingual RNN-based translation model, repurposing the source-side embeddings to define a similarity measure between multilingual sentence representations;
	\item \LaBSE: Similar to \base, LaBSE \citep{feng-etal-2022-language} uses a dual encoder model based on the BERT base architecture and is trained with a contrastive loss; it additionally benefits from pre-training on large sets of monolingual data, and covers more than 110 languages;
	\item \ft\rtvr{[model]-[Err]}:  we fine-tune either one of the aforementioned encoders with the losses of \textsection{\ref{ssec:training}}; Err is one of $\{\text{MSE}, \text{MAE}, \text{Rank}\}$ (e.g. \ft\LaBSE\MSE).
\end{itemize}

In our setting, fine-tuning updates all the parameters. The global learning rate is 1e-4; as for $a$ and $b$ (eq.~\ref{eq:mapping-function-cos-to-lev}) it is set to \text{1e-2}. Each source $\src$ is presented with three $\tilde \tgt$ (eq.~\ref{eq:loss-metric-learning}) determined by the top 3 scoring matches obtained with \basebow. The validation score is an in-batch normalized discounted cumulative gain (NDCG) score \citep{wang-etal-2013-theoretical}.
% ls=0.1
% validation score = in-batch NDCG
% lr=1e-4
% a,b lr=1e-2
% each src has 3 examples

% Retrieval of the most similar examples is implemented with the FAISS library\footnote{\url{https://faiss.ai/.}} \citep{douze-etal-2024-faiss} in an in-domain setting.
Retrieval is always performed in an in-domain setting, ensuring that translations for a specific domain are exclusively based on examples from that domain.
For all methods, a threshold is further applied to filter out low similarity segments. For each retrieval method, we adjust this threshold on the validation set to achieve an averaged retrieval rate\footnote{Fraction of sentences with at least one retrieved example.} of $50$\%. Using a constant retrieval rate for all retrievers ensures that no system is unduly rewarded for retrieving a greater number of similar examples, so that our comparisons of retrievers \emph{only reflect differences in the quality of the TM matches}.
% comparable This is important}
% ; a low threshold to have retrieval rates $>90\%$.
Refer to appendix~\ref{appendix:search} for a discussion of search configurations.

\subsection{NMT Architectures}

We consider three RANMT architectures. The first is an in-house implementation of \NFA \citep{bulte-tezcan-2019-neural}, which extends a basic encoder-decoder MT architecture by prepending the target side of TM examples to the source text on the encoder side. As is standard, the retrieved examples used in training are selected based on their Levenshtein distance to the source text; at most one similar example is used. This model thus also has the ability to perform inference from scratch, without using any retrieved example.

The second architecture implements an edit-based, non-autoregressive encoder-decoder model that has the ability to jointly edit up to $k$ examples to generate a translation, reproducing the multi-Levenshtein transformer (\mlevt{k}) \citep{bouthors-etal-2023-towards}. Retrieved examples used in training are also selected based on source-side similarity measures; we use up to $k=3$ examples.

The third architecture is \EuroLLM-9B \citep{martins-etal-2025-eurollm9b}, specifically designed and trained for multilingual tasks such as Machine Translation. We fine-tune the \EuroLLM-9B base model on the en-fr and de-en training corpora using a prompt \emph{that only contains relevant examples in the target language}. A discussion of the adaption of \EuroLLM for this setup is in Appendix~\ref{appendix:eurollm}.

\section{Results \label{sec:results}}

We first run controlled experiments serving two main purposes: (a) to compare the quality of various retrieval models (\textsection~\ref{ssec:clir_results}); (b) to evaluate the gap between the baseline source-side retrieval and our proposed CLIR-based approaches (\textsection~\ref{ssec:trans_results}), an analysis that crucially requires parallel datasets. These experiments use two language pairs (en-fr and de-en).\fydone{Keep ? add one more language ?} This configuration is somewhat ideal, as CLIR methods are primarily meant for cases where only monolingual data are available, as illustrated by the experiments in \textsection~\ref{ssec:mono_results}.

\begin{table}[t]
	\centering
	\resizebox{\linewidth}{!}{
		\begin{tabular}{lrrrr}
			% \begin{tabular}{l|rr|rr}
			\toprule
			                         & \multicolumn{2}{c}{test en-fr}                & \multicolumn{2}{c}{test de-en}                                                                                                                                        \\[-1pt]
			\cmidrule(lr){2-3}\cmidrule(lr){4-5}
			                         & \sm{\makebox[\widthof{XXX}]{\Lev~$\uparrow$}} & \sm{\makebox[\widthof{XXXX}]{\texttt{xsim}~$\downarrow$}} & \sm{\makebox[\widthof{XXX}]{\Lev~$\uparrow$}} & \sm{\makebox[\widthof{XXXX}]{\texttt{xsim}~$\downarrow$}} \\[-0.5pt] \midrule[0.75pt]
			\sm{\fuzzygold (oracle)} & 50.0                                          & -                                                         & 53.4                                          & -                                                         \\[-0.5pt]
			\sm{\fuzzysrc}           & 36.0                                          & -                                                         & 21.9                                          & -                                                         \\[-0.5pt]
			\sm{\fuzzybt}            & 31.1                                          & -                                                         & 21.8                                          & -                                                         \\[-0.5pt] \hline
			\sm{\base}               & 30.4                                          & 10.8                                                      & -                                             & -                                                         \\[-0.5pt]
			\sm{\basebow}            & 32.7                                          & 8.0                                                       & 30.9                                          & 25.2                                                      \\[-0.5pt] \hline
			\sm{\ft\basebow\MSE}     & 34.9                                          & 7.6                                                       & 33.6                                          & 25.2                                                      \\[-0.5pt]
			\sm{\ft\basebow\MAE}     & 35.2                                          & 8.0                                                       & 32.9                                          & 25.3                                                      \\[-0.5pt]
			\sm{\ft\basebow\Rank}    & 29.9                                          & 11.9                                                      & 25.9                                          & 41.5                                                      \\[-0.5pt] \hline
			\sm{\LASER}              & 32.8                                          & 10.0                                                      & 33.2                                          & 26.2                                                      \\[-0.5pt]
			\sm{\LaBSE}              & 33.8                                          & 7.4                                                       & 34.4                                          & \textbf{21.0}                                             \\[-0.5pt] \hline
			\sm{\ft\LaBSE\MSE}       & 36.7                                          & 7.6                                                       & 34.0                                          & 24.6                                                      \\[-0.5pt]
			\sm{\ft\LaBSE\MAE}       & \textbf{37.0}                                 & \textbf{6.4}                                              & \textbf{34.5}                                 & 23.5                                                      \\[-0.5pt]
			\sm{\ft\LaBSE\Rank}      & 36.0                                          & 7.3                                                       & 34.2                                          & 23.3                                                      \\[-0.5pt]
			\bottomrule
		\end{tabular}
	}
	\caption{\label{tab:xsim-all-domain} %\texttt{xsim} error rate and average Levenshtein similarity (1-best example to reference)
		Average retrieval scores (\texttt{xsim} error and Levenshtein similarity), averaged accross domains.}
\end{table}

\subsection{Retrieval Scores \label{ssec:clir_results}}

Table~\ref{tab:xsim-all-domain} stores the retrieval scores (\textsection{\ref{ssec:metrics}}) for the small-scale experiments.
% \tomodif{Overall, cross-lingual pre-trained models yield much better target side \Lev{} similarity values than source-side fuzzy matching}.
For en-fr, source-side fuzzy matching yields more similar segments than CLIR techniques, a gap that vanishes after fine-tuning with MAE for \basebow{} and even more so for \LaBSE. For de-en, the gap between the source-side fuzzy matching baseline and the oracle condition is much larger than for en-fr, which might be due to residual misalignments or to morphological differences between German and English.\footnote{Inflectional processes in German create multiple variants for each lexeme, making exact matches at the word level much less likely in German than in English.} All CLIR techniques widely outperform the monolingual settings (\fuzzysrc and \fuzzybt). For this language pair, however we do not see much gain from fine-tuning, perhaps owing to a smaller fune-tuning dataset.
% , we observe small additional gains from fine-tuning, that do not exist for de-en, where the parallel data available for fine-tuning is much smaller.
As expected, fine-tuning with a lexical loss is always slightly detrimental to the sentence-level retrieval scores (\texttt{xsim} error) for both \basebow{} and \LaBSE.
Training \basebow{} along with a bag-of-word loss (eq.~\eqref{eq:bow-loss}) effectively improves the retrieval scores.
% \todo{consistency in notations - bold or not }

\subsection{Translation Scores \label{ssec:trans_results}}

\paragraph{Retrieval-free Baselines} We use NLLB, \NFA and \EuroLLM without any retrieved examples as machine translation baselines. Their corresponding BLEU scores, averaged across domains, are respectively 39.0 (en-fr) and 33.8  (de-en) for NLLB and 48.0 (en-fr) and 38.2 (de-en) for \NFA. Details per domain are in Table~\ref{tab:bleu-per-domain-nfa-clir}.

\begin{table*}[t]
	\centering
	\resizebox{0.9\textwidth}{!}{
		\begin{tabular}{lrrrrrrrrrrrr}
			% \begin{tabular}{l|rr|rr}
			\toprule
			                         & \multicolumn{4}{c}{\mlevt{3}}  & \multicolumn{4}{c}{\NFA}             & \multicolumn{4}{c}{\EuroLLM}                                                                                                                                                                                                                                                                                         \\
			\cmidrule(lr){2-5}\cmidrule(lr){6-9}\cmidrule(lr){10-13}
			                         & \multicolumn{2}{c}{test en-fr} & \multicolumn{2}{c}{test de-en}       & \multicolumn{2}{c}{test en-fr} & \multicolumn{2}{c}{test de-en}       & \multicolumn{2}{c}{test en-fr} & \multicolumn{2}{c}{test de-en}                                                                                                                                                                              \\
			\cmidrule(lr){2-3}\cmidrule(lr){4-5}\cmidrule(lr){6-7}\cmidrule(lr){8-9}\cmidrule(lr){10-11}\cmidrule(lr){12-13}
			                         & \sm{BLEU}                      & \sm{\makebox[\widthof{BLEU}]{COMET}} & \sm{BLEU}                      & \sm{\makebox[\widthof{BLEU}]{COMET}} & \sm{BLEU}                      & \sm{\makebox[\widthof{BLEU}]{COMET}} & \sm{BLEU}           & \sm{\makebox[\widthof{BLEU}]{COMET}} & \sm{BLEU}     & \sm{\makebox[\widthof{BLEU}]{COMET}} & \sm{BLEU} & \sm{\makebox[\widthof{BLEU}]{COMET}} \\ \midrule[0.75pt]
			no example               & -                              & -                                    & -                              & -                                    & 48.0                           & 87.1                                 & 38.2                & 79.8                                 & 44.4          & 86.3                                 & 40.7      & 82.6                                 \\
			\sm{\fuzzygold (oracle)} & 45.3                           & 51.7                                 & 31.3                           & -8.6                                 & 52.2                           & 87.8                                 & 44.7                & 82.2                                 & 52.6          & 88.1                                 & 46.8      & 83.9                                 \\
			\sm{\fuzzysrc}           & \textbf{43.8}                  & \textbf{48.4}                        & 28.8                           & -10.3                                & \textbf{51.3}                  & \textbf{87.5}                        & 41.9                & \textbf{81.7}                        & \textbf{51.6} & \textbf{87.8}                        & 41.6      & 83.0                                 \\
			\sm{\fuzzybt}            & 40.5                           & 43.8                                 & 22.0                           & -22.9                                & 49.9                           & 87.2                                 & 38.9                & 81.3                                 & 46.4          & 86.9                                 & 41.8      & 82.9                                 \\ \cmidrule(lr){2-5}\cmidrule(lr){6-9}\cmidrule(lr){10-13}
			\sm{\base}               & 41.5                           & 42.2                                 & -                              & -                                    & 49.9                           & 87.1                                 & -                   & -                                    & 50.4          & 87.4                                 & -         & -                                    \\
			\sm{\basebow}            & 42.4                           & 44.1                                 & 28.3                           & -11.6                                & 50.5                           & 87.3                                 & 40.0                & 80.9                                 & 51.1          & 87.5                                 & 44.1      & 83.0                                 \\ \cmidrule(lr){2-5}\cmidrule(lr){6-9}\cmidrule(lr){10-13}
			\sm{\ft\basebow\MSE}     & 43.1                           & 46.0                                 & 29.0                           & -12.7                                & 51.0                           & \textbf{87.5}                        & 41.6                & 81.5                                 & 51.3          & \textbf{87.8}                        & 45.1      & \bf 83.3                             \\
			\sm{\ft\basebow\MAE}     & 43.2                           & 46.3                                 & 29.1                           & -12.6                                & 51.0                           & 87.4                                 & 41.8                & 81.4                                 & 51.5          & 87.7                                 & \bf 45.2  & 83.4                                 \\
			\sm{\ft\basebow\Rank}    & 42.1                           & 44.3                                 & 26.0                           & -17.9                                & 50.7                           & 87.3                                 & 41.8                & 80.9                                 & 50.4          & 87.5                                 & 43.6      & 83.1                                 \\ \cmidrule(lr){2-5}\cmidrule(lr){6-9}\cmidrule(lr){10-13}
			\sm{\LASER}              & 42.5                           & 44.3                                 & \textbf{30.4}                  & \textbf{-7.7}                        & 49.9                           & 87.3                                 & 41.1                & 81.3                                 & 50.9          & 87.4                                 & 44.9      & 83.1                                 \\
			\sm{\LaBSE}              & 43.1                           & 45.5                                 & 29.8                           & -9.5                                 & 50.5                           & 87.3                                 & 41.5                & 81.4                                 & 51.5          & 87.6                                 & 45.1      & 83.2                                 \\ \cmidrule(lr){2-5}\cmidrule(lr){6-9}\cmidrule(lr){10-13}
			\sm{\ft\LaBSE\MSE}       & \Spval\textbf{43.7}            & \Spval\textbf{47.8}                  & 27.7                           & -12.5                                & \Spval\textbf{51.2}            & \Spval\textbf{87.5}                  & 41.6                & \Spval\textbf{81.6}                  & \textbf{51.6} & \textbf{87.8}                        & 44.2      & 83.2                                 \\
			\sm{\ft\LaBSE\MAE}       & \Spval43.6                     & \Spval47.7                           & 29.1                           & -9.2                                 & \Spval\textbf{51.2}            & \Spval\textbf{87.5}                  & \Spval\textbf{42.0} & 81.5                                 & 51.5          & \textbf{87.8}                        & 44.9      & \bf 83.3                             \\
			\sm{\ft\LaBSE\Rank}      & \Spval43.3                     & \Spval47.1                           & 28.4                           & -12.2                                & \Spval51.1                     & \Spval\textbf{87.5}                  & \Spval41.8          & \Spval\textbf{81.6}                  & 51.3          & 87.7                                 & 44.8      & 83.2                                 \\ \bottomrule
		\end{tabular}
	}
	\caption{\label{tab:bleu-mlevt-nfa-auto-p-0.5}Average translation scores: \mlevt{3}, \NFA, \EuroLLM, all domains. Significant results (p=0.05) w.r.t. \LaBSE{} are marked with \Spval. Best CLIR results are in \textbf{bold}. When \fuzzysrc{} is not outperformed, its score is in \textbf{bold}.}
\end{table*}

\begin{table*}[ht]
	\centering
	\small
	% \begin{tabular}{|l||rrrrrrrrrrr||rrrrr|}
	\resizebox{\textwidth}{!}{
		\begin{tabular}{lrrrrrrrrrrrrrrrr}
			\toprule
			% & \multicolumn{11}{c||}{English-French} & \multicolumn{5}{c|}{German-English}                                                                                                                                                                                                                                                                                                                                                                                                                                                    \\
			                    & \multicolumn{11}{c}{English-French} & \multicolumn{5}{c}{German-English}                                                                                                                                                                                                                                                                                                                                                                                                                                                     \\[-1pt]
			\cmidrule(lr){2-12}\cmidrule(lr){13-17}
			~                   & \makebox[\widthof{xxx}]{ECB}        & \makebox[\widthof{xxx}]{EME}       & \makebox[\widthof{xxx}]{Epp} & \makebox[\widthof{xxx}]{GNO} & \makebox[\widthof{xxx}]{JRC} & \makebox[\widthof{xxx}]{KDE} & \makebox[\widthof{xxx}]{News} & \makebox[\widthof{xxx}]{PHP} & \makebox[\widthof{xxx}]{TED} & \makebox[\widthof{xxx}]{Ubu} & \makebox[\widthof{xxx}]{Wiki} & \makebox[\widthof{xxx}]{KDE} & \makebox[\widthof{xxx}]{Kor} & \makebox[\widthof{xxx}]{JRC} & \makebox[\widthof{xxx}]{EME} & \makebox[\widthof{xxx}]{Sub} \\[-2pt] \cmidrule(lr){1-12} \cmidrule(lr){13-17} \cmidrule(lr){1-12} \cmidrule(lr){13-17}
			\tn{\fuzzygold}     & 58.5                                & 64.8                               & 32.5                         & 61.3                         & 60.5                         & 44.3                         & 27.0                          & 41.0                         & 31.2                         & 40.5                         & 36.9                          & 32.1                         & 8.2                          & 48.7                         & 47.6                         & 19.8                         \\[-2pt]
			\tn{\fuzzysrc}      & 56.7                                & \bf 62.2                           & \bf 31.9                     & \bf 58.9                     & 58.8                         & 41.5                         & \bf 27.2                      & \bf 40.3                     & \bf 30.7                     & \bf 38.9                     & 34.6                          & 29.2                         & 8.7                          & 45.8                         & 43.3                         & 16.9                         \\[-2pt]
			\tn{\fuzzybt}       & \contamnllb{51.0}                   & \contamnllb{54.7}                  & \bf 31.9                     & 51.3                         & \contamnllb{55.1}            & 35.2                         & 27.0                          & 38.7                         & \bf 30.7                     & 36.2                         & 33.3                          & 21.3                         & 4.5                          & \contamnllb{27.5}            & \contamnllb{39.2}            & \contamnllb{\textbf{17.5}}   \\[-1pt] \cmidrule(lr){1-12} \cmidrule(lr){13-17}
			\tn{\LASER}         & 56.5                                & 60.5                               & 31.7                         & 56.0                         & 58.6                         & 36.6                         & 26.9                          & 39.7                         & \bf 30.6                     & 36.4                         & 33.9                          & \textbf{29.3}                & 11.3                         & 48.9                         & \textbf{46.6}                & 16.2                         \\[-2pt]
			\tn{\LaBSE}         & \bf 57.0                            & 59.8                               & 31.3                         & 57.8                         & \bf 59.4                     & 40.8                         & 26.9                          & 39.8                         & 30.5                         & 37.0                         & 33.9                          & 29.0                         & 9.3                          & \textbf{49.2}                & 45.4                         & 16.2                         \\[-2pt] \cmidrule(lr){1-12} \cmidrule(lr){13-17}
			\tn{\ft\LaBSE\MSE } & \bf 57.0                            & \bf 62.0                           & \bf 31.8                     & 58.6                         & 58.6                         & \bf 42.5                     & \bf 27.0                      & 39.9                         & 30.5                         & \bf 38.1                     & \bf 34.9                      & 28.1                         & 10.9                         & 41.4                         & 41.1                         & 17.0                         \\[-2pt]
			\tn{\ft\LaBSE\MAE } & 56.9                                & 61.4                               & 31.6                         & \bf 58.7                     & 58.7                         & \bf 42.5                     & \bf 27.0                      & 39.9                         & 30.4                         & 37.7                         & 34.5                          & 28.7                         & \textbf{11.7}                & 44.4                         & 43.7                         & 17.0                         \\[-2pt]
			\tn{\ft\LaBSE\Rank} & 56.6                                & 61.1                               & 31.7                         & 57.7                         & 58.0                         & 41.7                         & 26.9                          & \bf 40.1                     & 30.6                         & 37.6                         & 34.1                          & 28.7                         & 9.6                          & 44.6                         & 42.5                         & 16.6                         \\[-1pt]
			\bottomrule
		\end{tabular}
	}
	\caption{\label{tab:bleu-per-domain-mlevt-clir} Per domain \BLEU{} scores for \mlevt{3} systems. Contaminated domains for NLLB (used in back-translation) are \contamnllb{underlined}.}
\end{table*}

\begin{table*}[t]
	\centering
	\small
	\resizebox{\textwidth}{!}{
		% \begin{tabular}{|l||rrrrrrrrrrr||rrrrr|}
		\begin{tabular}{lrrrrrrrrrrrrrrrr}
			\toprule
			% & \multicolumn{11}{c||}{English-French} & \multicolumn{5}{c|}{German-English}                                                                                                                                                                                                                                                                                                                                                                                                                                                    \\
			                    & \multicolumn{11}{c}{English-French} & \multicolumn{5}{c}{German-English}                                                                                                                                                                                                                                                                                                                                                                                                                                                     \\[-1pt]
			\cmidrule(lr){2-12}\cmidrule(lr){13-17}
			~                   & \makebox[\widthof{xxx}]{ECB}        & \makebox[\widthof{xxx}]{EME}       & \makebox[\widthof{xxx}]{Epp} & \makebox[\widthof{xxx}]{GNO} & \makebox[\widthof{xxx}]{JRC} & \makebox[\widthof{xxx}]{KDE} & \makebox[\widthof{xxx}]{News} & \makebox[\widthof{xxx}]{PHP} & \makebox[\widthof{xxx}]{TED} & \makebox[\widthof{xxx}]{Ubu} & \makebox[\widthof{xxx}]{Wiki} & \makebox[\widthof{xxx}]{KDE} & \makebox[\widthof{xxx}]{Kor} & \makebox[\widthof{xxx}]{JRC} & \makebox[\widthof{xxx}]{EME} & \makebox[\widthof{xxx}]{Sub} \\[-2pt]
			\cmidrule(lr){2-12}\cmidrule(lr){13-17}
			contam. rate        & 33.9                                & 58.3                               & 8.9                          & 0                            & 2.1                          & 6.8                          & 100                           & 6.8                          & 29.8                         & 0                            & 0                             & 47.5                         & 0                            & 54.0                         & 17.3                         & 8.9                          \\[-2pt] \cmidrule(lr){1-12} \cmidrule(lr){13-17} \cmidrule(lr){1-12} \cmidrule(lr){13-17}
			\sm{\texttt{NLLB}}  & 44.5                                & 40.3                               & 35.9                         & 39.6                         & 50.7                         & 34.2                         & 32.1                          & 39.0                         & 41.0                         & 35.9                         & 35.4                          & 29.1                         & 23.5                         & 46.4                         & 38.8                         & 31.4                         \\[-2pt]
			\sm{no example}     & 58.9                                & 55.9                               & 39.9                         & 48.8                         & 62.3                         & 42.8                         & 50.1                          & 45.6                         & 43.9                         & 43.5                         & 36.4                          & 39.5                         & 13.2                         & 55.5                         & 51.6                         & 31.0                         \\[-2pt] \cmidrule(lr){1-12} \cmidrule(lr){13-17}
			\tn{\fuzzygold}     & 65.2                                & 65.1                               & 40.3                         & 59.7                         & 68.1                         & 46.6                         & 50.1                          & 48.1                         & 44.1                         & 47.8                         & 39.0                          & 45.6                         & 21.9                         & 63.9                         & 60.5                         & 31.7                         \\[-2pt]
			% \fuzzysrc (old) & 59.3                                & 57.6                               & 40.1                         & 49.9                         & 62.4                         & 42.8                         & 50.1                          & 45.8                         & 43.6                         & 44.8                         & 36.9                          & 40.0                         & 13.2                         & 55.6                         & 52.4                         & 31.0                         \\[-2pt]
			\tn{\fuzzysrc}      & \textbf{64.5}                       & \textbf{63.0}                      & \textbf{40.0}                & 58.4                         & \textbf{66.6}                & 45.2                         & 50.0                          & 47.3                         & 44.0                         & \textbf{46.8}                & \textbf{38.3}                 & \textbf{44.3}                & 14.2                         & \textbf{62.3}                & 57.4                         & 31.1                         \\[-2pt]
			\tn{\fuzzybt}       & \contamnllb{62.4}                   & \contamnllb{60.4}                  & 39.8                         & 54.7                         & \contamnllb{64.9}            & 44.5                         & \textbf{50.1}                 & 46.8                         & 43.7                         & 44.7                         & 37.0                          & 40.1                         & 13.2                         & \contamnllb{55.6}            & \contamnllb{54.8}            & \contamnllb{31.1}            \\[-1pt] \cmidrule(lr){1-12} \cmidrule(lr){13-17}
			\tn{\LASER}         & 63.7                                & 61.6                               & 39.8                         & 56.8                         & 65.4                         & 44.0                         & 50.0                          & 46.8                         & \textbf{44.1}                & 46.1                         & 37.1                          & 41.4                         & 16.1                         & 60.7                         & 56.0                         & 31.1                         \\[-2pt]
			\tn{\LaBSE}         & 63.8                                & 62.5                               & 39.0                         & 57.7                         & 65.0                         & 45.3                         & 50.0                          & 46.9                         & \textbf{44.1}                & 46.0                         & 37.7                          & 43.0                         & 15.6                         & 60.6                         & 56.9                         & 31.1                         \\[-2pt] \cmidrule(lr){1-12} \cmidrule(lr){13-17}
			\tn{\ft\LaBSE\MSE}  & \textbf{64.3}                       & \textbf{63.0}                      & \textbf{40.0}                & 58.2                         & \textbf{66.6}                & 45.3                         & \textbf{50.1}                 & \textbf{47.5}                & 44.0                         & \textbf{46.8}                & 37.9                          & 43.1                         & 16.8                         & 60.3                         & 56.9                         & \textbf{31.2}                \\[-2pt]
			\tn{\ft\LaBSE\MAE}  & 64.2                                & \textbf{63.0}                      & 39.9                         & \textbf{58.5}                & \textbf{66.6}                & \textbf{45.5}                & \textbf{50.1}                 & 47.4                         & 43.9                         & 46.3                         & 37.8                          & 43.3                         & \textbf{17.1}                & \textbf{61.4}                & \textbf{57.5}                & 30.9                         \\[-2pt]
			\tn{\ft\LaBSE\Rank} & 64.1                                & 62.6                               & 39.8                         & 58.1                         & 66.3                         & 45.3                         & 50.0                          & \textbf{47.5}                & 44.0                         & 46.0                         & \textbf{38.1}                 & \textbf{43.7}                & 15.9                         & 61.0                         & 57.2                         & \textbf{31.2}                \\[-1pt]
			\bottomrule
		\end{tabular}
	}
	\caption{\label{tab:bleu-per-domain-nfa-clir} Per domain \BLEU{} scores for NLLB and \NFA systems. Contaminated domains for NLLB (used in back-translation) are \contamnllb{underlined}. The contamination rate (top line) for \NFA{} is the percentage of overlap of its training data with the test set.}
\end{table*}

\begin{table*}[ht]
	\centering
	\small
	% \begin{tabular}{|l||rrrrrrrrrrr||rrrrr|}
	\resizebox{\textwidth}{!}{
		\begin{tabular}{lrrrrrrrrrrrrrrrr}
			\toprule
			% & \multicolumn{11}{c||}{English-French} & \multicolumn{5}{c|}{German-English}                                                                                                                                                                                                                                                                                                                                                                                                                                                    \\
			                    & \multicolumn{11}{c}{English-French} & \multicolumn{5}{c}{German-English}                                                                                                                                                                                                                                                                                                                                                                                                                                                     \\[-1pt]
			\cmidrule(lr){2-12}\cmidrule(lr){13-17}
			~                   & \makebox[\widthof{xxx}]{ECB}        & \makebox[\widthof{xxx}]{EME}       & \makebox[\widthof{xxx}]{Epp} & \makebox[\widthof{xxx}]{GNO} & \makebox[\widthof{xxx}]{JRC} & \makebox[\widthof{xxx}]{KDE} & \makebox[\widthof{xxx}]{News} & \makebox[\widthof{xxx}]{PHP} & \makebox[\widthof{xxx}]{TED} & \makebox[\widthof{xxx}]{Ubu} & \makebox[\widthof{xxx}]{Wiki} & \makebox[\widthof{xxx}]{KDE} & \makebox[\widthof{xxx}]{Kor} & \makebox[\widthof{xxx}]{JRC} & \makebox[\widthof{xxx}]{EME} & \makebox[\widthof{xxx}]{Sub} \\[-2pt] \cmidrule(lr){1-12} \cmidrule(lr){13-17} \cmidrule(lr){1-12} \cmidrule(lr){13-17}
			\tn{no example}     & 49.9                                & 47.7                               & 39.2                         & 45.7                         & 55.5                         & 40.9                         & 34.0                          & 48.1                         & 42.0                         & 41.3                         & 44.2                          & 41.6                         & 23.3                         & 54.4                         & 52.9                         & 31.1                         \\[-2pt]
			\tn{\fuzzygold}     & 63.7                                & 68.4                               & 39.5                         & 65.2                         & 66.2                         & 51.1                         & 33.8                          & 52.1                         & 41.9                         & 48.9                         & 48.2                          & 50.4                         & 25.6                         & 63.6                         & 62.2                         & 32.2                         \\[-2pt]
			\tn{\fuzzysrc}      & 62.6                                & \bf 66.9                           & \bf 39.5                     & \bf 63.3                     & 64.9                         & 49.0                         & 33.7                          & 51.6                         & \bf 41.7                     & \bf 47.4                     & 47.0                          & 43.3                         & 23.6                         & 55.1                         & 54.3                         & 31.8                         \\[-2pt]
			\tn{\fuzzybt}       & 52.0                                & 54.1                               & 39.3                         & 49.1                         & 56.5                         & 43.2                         & \bf 33.8                      & 50.4                         & 41.5                         & 45.0                         & 46.1                          & 42.5                         & 23.6                         & 54.8                         & 56.4                         & \bf 31.9                     \\[-1pt] \cmidrule(lr){1-12} \cmidrule(lr){13-17}
			\tn{\LASER}         & 63.2                                & 65.3                               & 39.4                         & 62.1                         & 65.5                         & 46.1                         & \bf 33.8                      & 51.0                         & \bf 41.7                     & 45.8                         & 46.4                          & 46.6                         & \bf 24.4                     & 62.4                         & 59.9                         & 31.4                         \\[-2pt]
			\tn{\LaBSE}         & \bf 63.5                            & 66.1                               & 39.4                         & \bf 63.3                     & \bf 65.8                     & 48.4                         & \bf 33.8                      & 51.2                         & 41.6                         & 46.4                         & 46.5                          & 47.2                         & 23.7                         & \bf 62.7                     & \bf 60.3                     & 31.4                         \\[-2pt] \cmidrule(lr){1-12} \cmidrule(lr){13-17}
			\tn{\ft\LaBSE\MSE } & 62.8                                & 66.3                               & 39.4                         & 63.2                         & 64.9                         & \bf 50.0                     & \bf 33.8                      & \bf 51.7                     & \bf 41.7                     & 46.8                         & \bf 47.3                      & 46.6                         & 24.1                         & 60.0                         & 58.7                         & 31.5                         \\[-2pt]
			\tn{\ft\LaBSE\MAE } & 62.8                                & 65.9                               & 39.4                         & \bf 63.3                     & 65.1                         & 49.6                         & \bf 33.8                      & 51.5                         & \bf 41.7                     & 46.8                         & 47.1                          & \bf 47.6                     & 24.2                         & 61.3                         & 59.9                         & 31.6                         \\[-2pt]
			\tn{\ft\LaBSE\Rank} & 62.4                                & 65.3                               & 39.4                         & 62.7                         & 64.6                         & 49.3                         & \bf 33.8                      & 51.5                         & \bf 41.7                     & \bf 47.3                     & 46.7                          & 47.5                         & 23.8                         & 61.1                         & 60.2                         & 31.5                         \\[-1pt]
			\bottomrule
		\end{tabular}
	}
	\caption{\label{tab:bleu-per-domain-eurollm-clir} Per domain \BLEU{} scores for \EuroLLM.}
\end{table*}

\paragraph{\mlevt{3}} Translation results for \mlevt{3}
% for all retrievers with an automatic threshold\todo{clarify}
are in Table~\ref{tab:bleu-mlevt-nfa-auto-p-0.5} (left part). The difference between the oracle (\fuzzygold) and the baseline (\fuzzysrc) gives an idea of the quality of the monolingual matches: the quality is pretty high in en-fr, owing to the cleaning procedure used for these data, more noisy in de-en, with a gap of about $2.5$ BLEU points. In comparison, using artificial back-translated source queries performs much worse than using natural texts. This difference is much larger than what was expected from the retrieval scores of Table~\ref{tab:xsim-all-domain}.

For both language pairs, using cross-lingual search enables to obtain results that compare to source-side fuzzy matching. For en-fr, the best scores in our setting use fine-tuned versions of \LaBSE, closing the gap in BLEU with \fuzzysrc, and delivering a small \COMET gain. Similar to the retrieval scores (Table~\ref{tab:xsim-all-domain}), we do not observe such benefit of fine-tuning for de-en, where cross-lingual retrieval with \LASER yields the best overall results, yielding a $1.6$ BLEU points improvement over the pure source-side matching.
% fine-tuning with MAE outperforms for en-fr (COMET).
% \fytodo{Check this - all domains? Also: the lexical gain is small}

\paragraph{\NFA} The corresponding results for \NFA{} are in Table~\ref{tab:bleu-mlevt-nfa-auto-p-0.5} (middle part). This model is a much stronger baseline than \mlevt{3} and the progress margins (with respect to \fuzzygold) is dimmer. Yet, we see that (a) retrieving examples directly on the target side (with \base, \LASER, and \LaBSE) improves the example-free baseline by about $3$ \BLEU{} points; (b) further fine-tuning the cross-lingual retrievers (with MSE and MAE) fully closes the gap with source-side matching for both metrics and language pairs. We observe that back-translating the monolingual target data also improves baseline systems, albeit by a much smaller margin.

% \paragraph{Source-side fuzzy matching vs. cross-lingual retrieval}
% We show that \fuzzysrc is a strong baseline with \LaBSE performing below its level. Leveraging the target side with back-translation (\fuzzybt) proves to be a rather ineffective approach.

% \maxTodo{Attendre résultats NFA avant d'en dire plus}
% \tomodif{Surprisingly, source-to-source fuzzy matching, which is the default setting for RANMT, obtains the worse overall results. Yet, the lexical nature of the retriever is not to blame, as the oracle target-to-target fuzzy matcher performs the best. This issue likely arises from the discrepancy between source and target language patterns, as well as a potential source-target parallel misalignments: source-side lexical similarity does not necessarily imply target side lexical similarity.
% % \fytodo{Comment : this is not likely given the efforts to clean data}
% Source-to-source retrieval with back-translation slightly mitigates this issue as they usually deliver more litteral source-target matchings. Similar observations about the benefits of data distillation \cite{kim-rush-2016-sequence} for NAR models are reported e.g., in \cite{zhou-2021-understanding}.}

\paragraph{\EuroLLM} Results for the fine-tuned version of \EuroLLM are in the right part of Table~\ref{tab:bleu-mlevt-nfa-auto-p-0.5}. This system turns out to be the most effective system overall, slightly outperforming NFA with $k=3$ TM examples. A first observation is that this model is particularly strong for translating into English (from German) where using a CLIR outperforms all baselines. For this model, the averaged gains achieved by fine-tuning the cross-lingual retrievers are very small, and vary accross domains (details in Table~\ref{tab:bleu-per-domain-eurollm-clir}). For this system, the gap between the \LaBSE{} system and the oracle condition (\fuzzygold) is already small in the two language pairs, showcasing the very strong built-in cross-lingual abilities of this model; yet, the difference that remains suggests that lexical similarity should still matter.
\fydone{Further comment}\fydone{k=3 may make a difference --> it doesnt}

\paragraph{Per-domain analysis} Per-domain BLEU scores are in Tables~\ref{tab:bleu-per-domain-mlevt-clir} (\mlevt{3}), \ref{tab:bleu-per-domain-nfa-clir} (\NFA), and \ref{tab:bleu-per-domain-eurollm-clir} (\EuroLLM). The domains for which we expect CLIR methods to work well need to satisfy two conditions: (a) a large enough set of examples to retrieve from, so that the choice the retriever actually makes a difference; (b) a small diversity of texts in that domain, so that retrieved instances are sufficiently similar to the input. This is well reflected for instance in Table~\ref{tab:bleu-per-domain-mlevt-clir} (\mlevt{3}) where CLIR techniques (with fine-tuning in en-fr, without in de-en) match the performance of the baseline in all domains, and even outpeform it in specialized domains such as ECB (finance), JRC (law), KDE (information technology) and Wikipedia. We do not see the same gains in Ubuntu (very small TM) or Europarl (large TM, with a diverse set of texts).

For \NFA, we also observe that using CLIR techniques yields scores that match the baseline; for both monolingual or cross-lingual retrieval, the improvement margin, as defined by the difference with \fuzzygold, is small for all domains, except for Koran (de-en), where CLIR techniques once again demonstrate their effectiveness.
% Some domains are harder to improve with our approach (e.g., News-Commentary or TED2013) -- which is expected, given the absence of gap between example-free and \fuzzygold scores for \NFA -- while others largely benefit from it. We observe the largest improvements for \mlevt{3} on EMEA, Gnome, KDE, Ubuntu and Wikipedia (all en-fr); while \NFA with cross-lingual retrieval has improved BLEU scores for Europarl (en-fr), Gnome, (en-fr), JRC-Acquis (de-en) and Koran (de-en).

Regarding \EuroLLM, we note that performing CLIR with the default version of \LaBSE{} already yields near optimal performance for many domains. Fine-tuning has here a reduced effect on \BLEU{} scores, sometimes slightly improving (e.g., for en-fr: EMEA, KDE, PHP, Wiki, UBU) or, more rarely, decreasing (e.g., for en-fr, ECB, JRC) performance. These variations highlight the fact that optimal translation scores ultimately also depend on the retrieval pool and the associated example quality. 

\paragraph{Comparison of fine-tuning losses}
The standard contrastive loss of eq.~\eqref{eq:contrastive-loss}, when applied in MT, primarily seeks to identify parallel sentences within collections of monolingual segments. Yet, RANMT requires examples that are lexically similar to the source sentences. The bag-of-word loss of eq.~\eqref{eq:bow-loss} effectively increases the rate of lexically similar sentences, thus enhancing the translation scores. As for our three fine-tuning strategies, \rtvr{MSE} and \rtvr{MAE} (eq.~\eqref{eq:loss-metric-learning}) yield comparable performances. The per-domain analysis further highlights their consistent behavior. As for \rtvr{Rank} (eq.~\eqref{eq:rankloss}), we observe that the rank-based loss consistently underperforms the other two, with a few exceptions (PHP for \NFA; PHP, Wiki and KDE (de-en) for \mlevt{3}).

\paragraph{On data contamination\label{par:data-contamination}}
We identify two sources of data contamination. First, \NFA training data partly overlaps with some of our test sets, in varying proportion (from 0 to 100) across domains (see Table~\ref{tab:bleu-per-domain-nfa-clir}). The intuition is that such contamination should make this model less sensitive to the choice of good examples. In the per-domain results of Table~\ref{tab:bleu-per-domain-nfa-clir}, this sensitivity is assessed by the gap between \rtvr{example-free} and \fuzzygold scores. As this gap is large for most domains (except
Europarl and News in en-fr, and Sub in de-en), we can still clearly assess the benefits of cross-lingual retrieval for the large majority of domains. Second, the training data of NLLB, used for back-translation, also partly overlaps with our test, which may boost the efficiency of the \fuzzybt baseline. Even for these domains (e.g., ECB, EMEA, JRC-Acquis), cross-lingual methods still achieve much higher BLEU scores than back-translation.

\begin{figure*}[ht]
	\centering
	\includegraphics[width=0.49\textwidth]{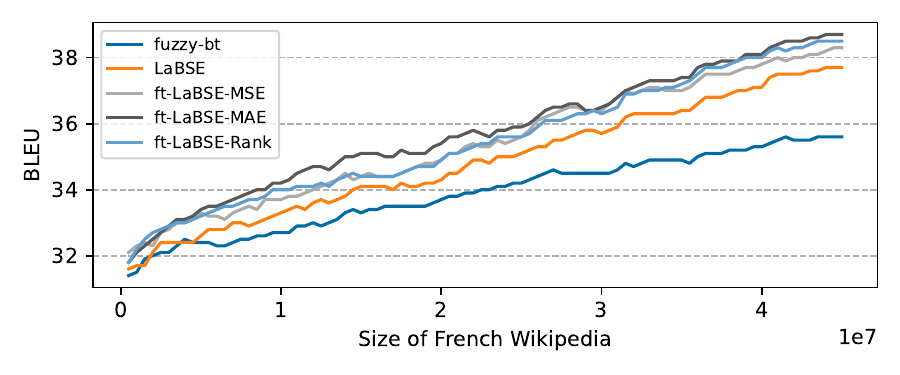}
	\includegraphics[width=0.49\textwidth]{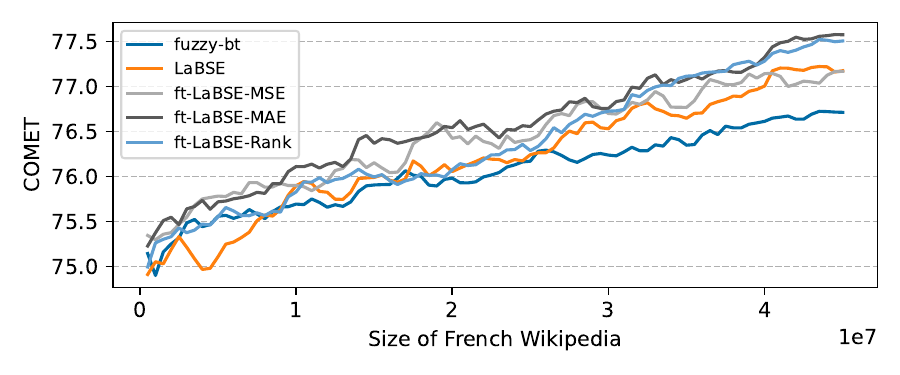}

	% \vspace{0pt}
	%   \caption{\label{fig:mlevt-bleu-comet-wikimono}Evolution of BLEU and COMET scores of \mlevt{3} with a growing retrieval pool.}
	% \end{figure*}

	% \begin{figure*}[ht]
	\centering
	\includegraphics[width=0.49\textwidth]{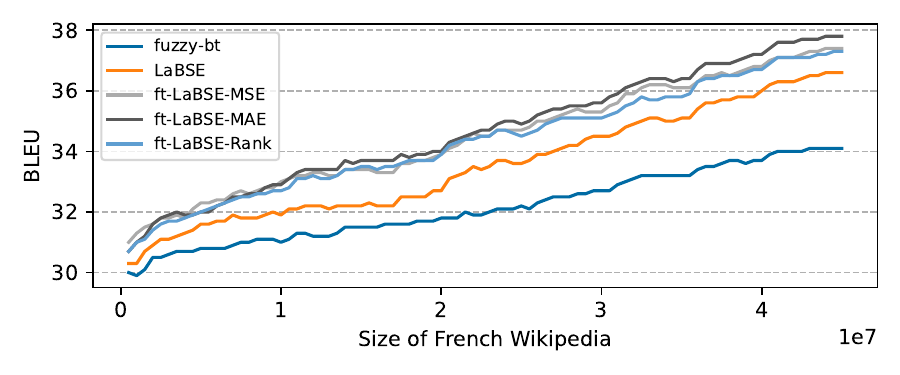}
	\includegraphics[width=0.49\textwidth]{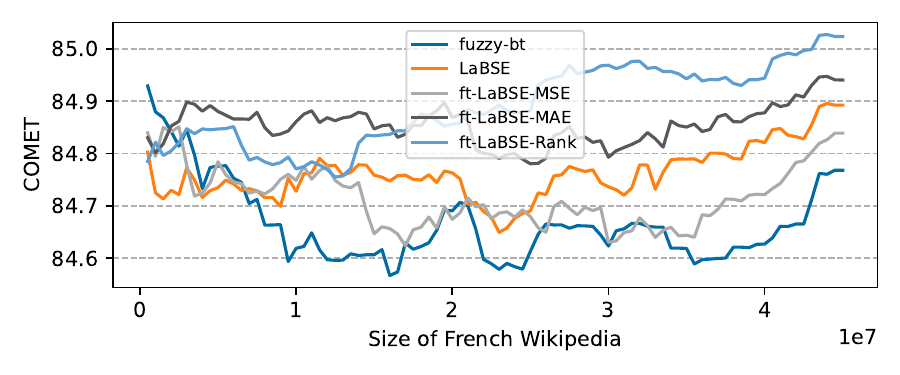}

	\centering
	\includegraphics[width=0.49\textwidth]{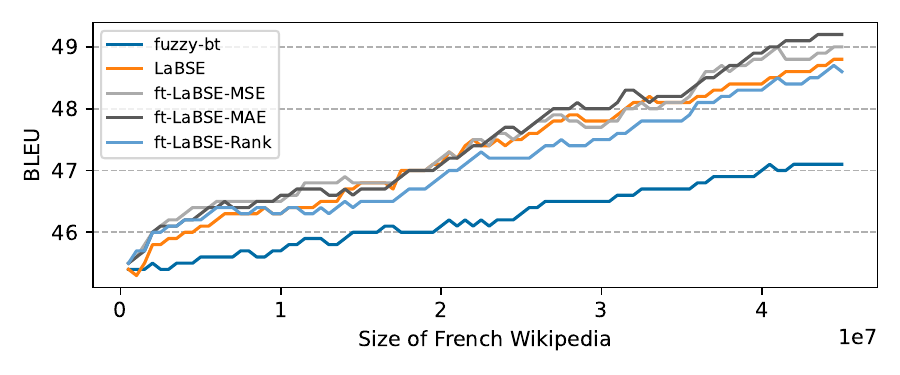}
	\includegraphics[width=0.49\textwidth]{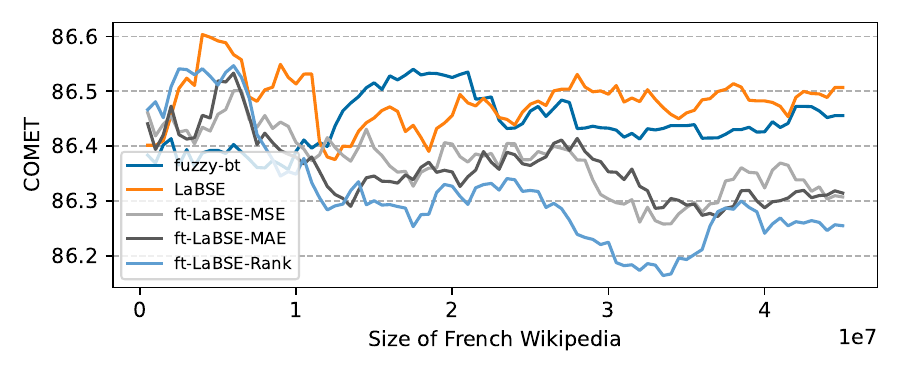}

	\caption{\label{fig:mlevt-nfa-bleu-comet-wikimono}\BLEU and \COMET scores with a growing retrieval pool: \mlevt{3} (top), \NFA (middle) and \EuroLLM (bottom).}
\end{figure*}

% \paragraph{Computational cost}
% % During training, retrieval is performed off-line once for each training instance, before training starts.
% For training, retrieval is performed offline, once for each source sentence, before the training process begins.
% Before inference, the TM is indexed for fuzzy matching and the target monolingual dataset is encoded once for each CLIR embbeder.
% During inference, running fuzzy matching retrievers only requires to tokenize the source sentence $\src$ and compute BM25 w.r.t. all examples in the TM, then \Lev{} on a small subset of the 100 most relevant segments.
% Regarding the cross-lingual dense retrievers, they must first encode $\src$, then perform retrieval with a $k$NN search.
% We compared both techniques on ECB (150K) with optimal configurations (1 CPU with 8 cores for fuzzy matching;\footnote{The search for the closest examples is multithreaded on the 8 cores.} a V100-32GB GPU with batching\footnote{The encoding step is performed in parallel for batches of $50$ segments. Parallelisation of search is handled by the Faiss library.} for CLIR). Each method can retrieve the 1st best match in $\sim{1}$ms on average.

% \cleardoublepage{}
\subsection{Large-scale Experiments \label{ssec:mono_results}}

\paragraph{Retrieving in a monolingual corpus} We use the models trained on the en-fr\footnote{A comparable experiment, targeting translation from English into Ukrainian is in Appendix~\ref{sec:en2uk}; these results also highlight the benefits of CLIR over monolingual retrieval in parallel data, especially when using a fine-tuned version of \LaBSE.} dataset to retrieve segments from the large monolingual Wikipedia dataset (see~\textsection{\ref{ssec:datasets}}), using the Wikipedia test already used in the controlled experiments set to assess performance. We use the same filtering threshold as for the controlled experiments to keep the settings comparable and ensure that the only difference is the increased size of the retrieval set. We only consider the variants of \LaBSE{}, which achieve the best results in the small data condition.
% Excluding \LASER from these experiments also spares the encoding of 45M segments: as explained in \textsection~\ref{ssec:retrieval-config}, LASER is based on recurrent networks and is much slower to run than the Translformer-based LaBSE.

Wikipedia is a challenging domain, due to the variety of topics that are addressed in the encyclopedia and the high level of formality of the text. NLLB already represents a strong baseline for that domain. Results are in Table~\ref{tab:bleu-comet-mlevt-nfa}. The first observation is that increasing the retrieval set (from 200K segments in the controlled experiments -- Tables~\ref{tab:bleu-per-domain-mlevt-clir} and~\ref{tab:bleu-per-domain-nfa-clir} -- to 45M segments) yields significant performance boosts across the board (e.g, +1.6 BLEU for \NFA and \fuzzybt, +0.4 BLEU points for \NFA and \LaBSE). With an additional lexical fine-tuning step, the benefits of the cross-lingual approach are again very clear, with large \BLEU{} differences relative to the small data condition:+3.9 for \ft\LaBSE\MAE with \mlevt{3}, +3.4 when using \NFA, +2 using \EuroLLM.

\begin{table}[t]
	\centering
	\resizebox{\linewidth}{!}{
		\begin{tabular}{lrrrrrr}
			\toprule
			                    & \multicolumn{2}{c}{\mlevt{3}} & \multicolumn{2}{c}{\NFA}             & \multicolumn{2}{c}{\EuroLLM}                                                                                               \\[-1pt]
			\cmidrule(lr){2-3}\cmidrule(lr){4-5}\cmidrule(lr){6-7}
			                    & \sm{BLEU}                     & \sm{\makebox[\widthof{BLEU}]{COMET}} & \sm{BLEU}                    & \sm{\makebox[\widthof{BLEU}]{COMET}} & \sm{BLEU}     & \sm{\makebox[\widthof{BLEU}]{COMET}} \\[-0.1pt] \midrule[0.75pt]
			no example          & -                             & -                                    & 36.4                         & 84.8                                 & 44.2          & 86.3                                 \\[-0.1pt] \hline
			\sm{\fuzzybt}       & 35.6                          & 76.7                                 & 38.6                         & 84.8                                 & 47.1          & 86.4                                 \\[-0.1pt] \hline
			\sm{\LaBSE}         & 37.7                          & 77.2                                 & 40.1                         & \textbf{85.0}                        & 48.8          & \textbf{86.5}                        \\[-0.1pt] \hline
			\sm{\ft\LaBSE\MSE}  & 38.3                          & 77.2                                 & 40.9                         & 84.8                                 & 49.0          & 86.3                                 \\[-0.1pt]
			\sm{\ft\LaBSE\MAE}  & \textbf{38.7}                 & \textbf{77.6}                        & \textbf{41.2}                & 84.9                                 & \textbf{49.2} & 86.3                                 \\[-0.1pt]
			\sm{\ft\LaBSE\Rank} & 38.5                          & 77.5                                 & 40.8                         & \textbf{85.0}                        & 48.6          & 86.3                                 \\[-0.1pt] \bottomrule
		\end{tabular}
	}
	\caption{\label{tab:bleu-comet-mlevt-nfa}\BLEU and \COMET scores with systems \mlevt{3}, \NFA and \EuroLLM with the Wikipedia retrieval pool.}
\end{table}

\paragraph{Effect of the retrieval pool size}
To better analyse the effect of the retrieval pool size, we progressively increase the Wikipedia retrieval set by batches of $500$K sentences from $500$K to $45$M, and compute the corresponding translation scores for each retriever.
% size of the $45$M sentences Wikipedia corpus. We retrieve the examples with various techniques in the first $N$ sentences of the corpus, where $N$'s are all multiples of $500$k.
Figure~\ref{fig:mlevt-nfa-bleu-comet-wikimono} displays the evolution of the translation metrics for \LaBSE, its fine-tuned variants and \fuzzybt. Overall, for all models, the benefit of growing the retrieval pool yields an almost linear \BLEU{} increase. \COMET variations are much smaller, especially for \NFA and \EuroLLM where it remains nearly constant. For both metrics, \ft\LaBSE\MAE outperforms the other setups for all data sizes in most cases.
% \fuzzybt initially obtains better BLEU scores than \LaBSE, but this difference vanishes as the retrieval pool increases.
\fuzzybt consistently lags behind CLIR alternatives, especially with \mlevt{3}.

% In figure~\ref{fig:best-of-each-bleu-comet-wikimono}, we compare the translation metrics of models \mlevt{3} and \NFA with \ft\LaBSE\MAE compared to the 1-best retrieved example.\footnote{1-best uses \ft\LaBSE\MSE which provides the highest scores among all.}
% The suprising conclusion is that both models degrade the BLEU score from $N\approx 34$M rather than simply copying the example. There is a decorellation between BLEU and COMET that we can possibly explain by a difference between lexical accuracy (BLEU) and fluency (COMET). \NFA is an autoregressive system that is here trained on a high quantity of high quality parallel data. It might favor translations lexically further away from the reference, but with higher general fluency, which is rewarded by a higher COMET score. There is a question about whether COMET is adapted to judge very domain-specific datasets that might have a low general fluency.

We also monitor in Table~\ref{tab:wiki-test-lev-rr} two additional scores: the retrieval rate (percentage of segments for which at least one example is found) and the average $\Lev$ similarity between the reference and its closest retrieved neighbor. For all retrieval methods, we see large increases, meaning a larger number of examples is retrieved and that their average quality (i.e, their similarity to the reference) also improves, highliting the usefulness of monolingual data for this domain.

\begin{table}[h]
	\centering
	\begin{tabular}{lrrrr}
		\toprule
		                    & \multicolumn{2}{c}{TM} & \multicolumn{2}{c}{mono}                                   \\
		\cmidrule(lr){2-3}\cmidrule(lr){4-5}
		                    & RR~$\uparrow$             & \Lev~$\uparrow$          & RR~$\uparrow$ & \Lev~$\uparrow$ \\ \midrule
		\sm{\fuzzybt}          & 39.9                   & 24.0                     & 60.2          & 31.0            \\
		\sm{\LaBSE}             & 30.9                   & 25.6                     & 59.6          & 35.2            \\
		\sm{\ft\LaBSE\MSE}  & 33.8                   & 27.2                     & 77.7          & 37.7            \\
		\sm{\ft\LaBSE\MAE}  & 31.3                   & 27.5                     & 59.0          & 38.1            \\
		\sm{\ft\LaBSE\Rank} & 25.7                   & 26.7                     & 61.2          & 36.7            \\
		\bottomrule
	\end{tabular}
	\caption{\label{tab:wiki-test-lev-rr}Comparison of the retrieval rates (RR) and average Levenshtein similarity ($\Lev$) on the Wikipedia test set of the systems retrieving in the TM vs. in the large monolingual corpus (mono).}
\end{table}

The translation examples in Appendix~\ref{sec:examples} illustrate the benefits of enlarging the retrieval pools, notably because they provide longer lexical matches, thereby improving the generated text.

% fuzzy-bt 0.4 rate = 0.602
% labse 0.71 rate = 0.596

% ft-labse-mse 0.85 rate = 0.777
% ft-labse-mae 0.9 rate = 0.59
% ft-labse-rank 0.55 rate = 0.612

% \todo{One / two examples, or an analysis of the number / quality of matches would help}

% \begin{figure*}[ht]
%   \centering
%   \includegraphics[width=0.49\textwidth]{clir/wikimono/curves-bleu-best-of-each.pdf}
%   \includegraphics[width=0.49\textwidth]{clir/wikimono/curves-comet-best-of-each.pdf}
%   \caption{\label{fig:best-of-each-bleu-comet-wikimono}Evolution of BLEU and COMET scores of the best of the three systems when growing the retrieval pool: \mlevt{3}, \NFA and 1-best.}
% \end{figure*}

% \paragraph{Effect of the threshold} We observed in our experiments the effect of the threshold. \NFA and \mlevt{3} systematically benefit from a tuned threshold.

% \maxTodo{rédiger}

\section{Conclusions and Outlook}

In this paper, we have explored various approaches to take advantage of monolingual examples in retrieval-augmented neural machine translation. Our main conclusion, resulting from experiments on 3 language directions and 16 varied textual domains, is that retrieving directly in the target language
% \tomodif{is preferable to performing fuzzy-matching on the source-side of back-translated data. This benefit is already clear with generic cross-lingual retrievers (LASER and LaBSE), and increases when the embeddings are further fine-tuned with dedicated losses targeting lexical matching objectives.}
can be as effective as fuzzy-matching on the source-side, especially with lexical-based fine-tuning objectives.
Using these techniques, we were also able to obtain large \BLEU{} increases over strong baselines on the Wikipedia dataset (up to +3.8 \BLEU{} points for our best configuration),
showcasing the benefits of RANMT in a large-scale setup. In our implementation, cross-lingual retrieval is no more costly than performing fuzzy matches in the source language.
% \todo{we may discuss this a bit more} %\maxTodo{Proof of concept $\rightarrow$ Labse should be fine-tuned on many languages}

% \todo{Implementation speed ?} \todo{50\% limit, } \todo{training \NFA with XL examples}

% \tomodif{While these results are already extremely promising}, there are several obvious ways to improve them:
We identify several ways to improve these results:
(a) by fine-tuning the retrieval models on very large multilingual datasets, adopting data settings that are comparable to the pre-training of LASER/LaBSE; (b) using more recent sentence embedders such as SONAR \citep{ambroise-etal-2023-sonar}; (c) using CLIR techniques to train the RANMT models, instead of relying on source-side matches as we have done here; (d) revisiting the other aspect of the retrieval pipeline -- e.g., relaxing the filtering threshold during example selection.

\section*{Limitations}
In this work, we have focused on extending a translation setting that is common in the translation industry, where translation memories have repeatedly been shown to speed up the translation process and improve the consistency of the resulting texts. In such setting, it is also common to have access to large monolingual resources, as has been studied eg., by \citep{tezcan-etal-2024-improving} for the legal domain and for \citep{tamura-etal-2023-target} in the scientific domain. By design, this setting is only applicable for so-called ``high-resource" language pairs, in which high-quality systems can be built and improved upon using auxiliary resources. In our selection of language pairs, we have favored domain diversity over language diversity, to better highlight how the availability of relevant examples impacts the overall translation quality. As our experiments show, the effectiveness of the cross-lingual approach does not depend on the similarity of the source and target languages, but mostly relies on the quality of cross-lingual sentence alignment. As shown e.g.,  by \citep{feng-etal-2022-language}, very precise alignments can be achieved, even for distant language pairs such as English-Chinese.
\fydone{Language diversity, scripts, etc}

We have thus deliberately chosen to exclude translation from and into ``low-resource'' languages, for which many resources and components are still difficult to collect (eg., parallel corpora to train sufficiently good NMT baselines, to be used for back translation; and also training and test data in specialized domains, where the TM-based approach is most likely to retrieve very good matches). We reckon that also improving generic NMT systems that could handle translations for those language pairs is another important research area, albeit distinct from what we have chosen to study in this work.

\section*{Ethical Statement}
There are no ethical issues.

% \appendix

\section*{Acknowledgments}

This research was funded by the French Agence Nationale de la Recherche (ANR) under the project TraLaLaM (“ANR-23-IAS1-0006”). This work was performed using HPC resources from GENCI–IDRIS (Grant 2024-A0161015117).

% GENCI IDRIS + Tralalam
% \clearpage{}

% \newpage

%% The file named.bst is a bibliography style file for BibTeX 0.99c
% \bibliographystyle{named}
\bibliography{../anthology,../custom}

\begin{thebibliography}{70}
\providecommand{\natexlab}[1]{#1}

\bibitem[{Agrawal et~al.(2023)Agrawal, Zhou, Lewis, Zettlemoyer, and Ghazvininejad}]{agrawal-etal-2023-context}
Sweta Agrawal, Chunting Zhou, Mike Lewis, Luke Zettlemoyer, and Marjan Ghazvininejad. 2023.
\newblock \href {https://doi.org/10.18653/v1/2023.findings-acl.564} {In-context examples selection for machine translation}.
\newblock In \emph{Findings of the Association for Computational Linguistics: ACL 2023}, pages 8857--8873, Toronto, Canada. Association for Computational Linguistics.

\bibitem[{Aharoni and Goldberg(2020)}]{aharoni-goldberg-2020-unsupervised}
Roee Aharoni and Yoav Goldberg. 2020.
\newblock \href {https://doi.org/10.18653/v1/2020.acl-main.692} {Unsupervised domain clusters in pretrained language models}.
\newblock In \emph{Proceedings of the 58th Annual Meeting of the Association for Computational Linguistics}, pages 7747--7763, Online. Association for Computational Linguistics.

\bibitem[{Artetxe and Schwenk(2019{\natexlab{a}})}]{artetxe-schwenk-2019-margin}
Mikel Artetxe and Holger Schwenk. 2019{\natexlab{a}}.
\newblock \href {https://doi.org/10.18653/v1/P19-1309} {Margin-based parallel corpus mining with multilingual sentence embeddings}.
\newblock In \emph{Proceedings of the 57th Annual Meeting of the Association for Computational Linguistics}, pages 3197--3203, Florence, Italy. Association for Computational Linguistics.

\bibitem[{Artetxe and Schwenk(2019{\natexlab{b}})}]{artetxe-schwenk-2019-massively}
Mikel Artetxe and Holger Schwenk. 2019{\natexlab{b}}.
\newblock \href {https://doi.org/10.1162/tacl_a_00288} {Massively multilingual sentence embeddings for zero-shot cross-lingual transfer and beyond}.
\newblock \emph{Transactions of the Association for Computational Linguistics}, 7:597--610.

\bibitem[{Arthern(1978)}]{arthern-1978-machine}
Peter~J. Arthern. 1978.
\newblock \href {https://aclanthology.org/1978.tc-1.5/} {Machine translation and computerised terminology systems - a translator`s viewpoint}.
\newblock In \emph{Translating and the Computer}, London, UK. Aslib Proceedings.

\bibitem[{Bawden and Yvon(2023)}]{bawden-yvon-2023-investigating}
Rachel Bawden and Fran{\c{c}}ois Yvon. 2023.
\newblock \href {https://aclanthology.org/2023.eamt-1.16/} {Investigating the translation performance of a large multilingual language model: the case of {BLOOM}}.
\newblock In \emph{Proceedings of the 24th Annual Conference of the European Association for Machine Translation}, pages 157--170, Tampere, Finland. European Association for Machine Translation.

\bibitem[{Bellet et~al.(2015)Bellet, Habrard, and Sebban}]{bellet-etal-2015-metric}
Aur{\'e}lien Bellet, Amaury Habrard, and Marc Sebban. 2015.
\newblock \emph{Metric learning}.
\newblock Morgan \& Claypool Publishers.

\bibitem[{Bogoychev and Sennrich(2020)}]{bogoychev-sennrich-2020-domaintranslationese}
Nikolay Bogoychev and Rico Sennrich. 2020.
\newblock \href {https://arxiv.org/abs/1911.03362} {Domain, translationese and noise in synthetic data for neural machine translation}.
\newblock \emph{CoRR}, abs/1911.03362.

\bibitem[{Bojar and Tamchyna(2011)}]{bojar-tamchyna-2011-improving}
Ond{\v{r}}ej Bojar and Ale{\v{s}} Tamchyna. 2011.
\newblock \href {https://aclanthology.org/W11-2138/} {Improving translation model by monolingual data}.
\newblock In \emph{Proceedings of the Sixth Workshop on Statistical Machine Translation}, pages 330--336, Edinburgh, Scotland. Association for Computational Linguistics.

\bibitem[{Bouthors et~al.(2023)Bouthors, Crego, and Yvon}]{bouthors-etal-2023-towards}
Maxime Bouthors, Josep Crego, and Fran{\c{c}}ois Yvon. 2023.
\newblock \href {https://doi.org/10.18653/v1/2023.emnlp-main.113} {Towards example-based {NMT} with multi-{L}evenshtein transformers}.
\newblock In \emph{Proceedings of the 2023 Conference on Empirical Methods in Natural Language Processing}, pages 1830--1846, Singapore. Association for Computational Linguistics.

\bibitem[{Bouthors et~al.(2024)Bouthors, Crego, and Yvon}]{bouthors-etal-2024-retrieving}
Maxime Bouthors, Josep Crego, and Fran{\c{c}}ois Yvon. 2024.
\newblock \href {https://doi.org/10.18653/v1/2024.findings-naacl.190} {Retrieving examples from memory for retrieval augmented neural machine translation: A systematic comparison}.
\newblock In \emph{Findings of the Association for Computational Linguistics: NAACL 2024}, pages 3022--3039, Mexico City, Mexico. Association for Computational Linguistics.

\bibitem[{Bowker(2002)}]{bowker-2002-computer}
Lynne Bowker. 2002.
\newblock \emph{Computer-aided translation technology: A practical introduction}.
\newblock University of Ottawa Press.

\bibitem[{Brown et~al.(2020)Brown, Mann, Ryder, Subbiah, Kaplan, Dhariwal, Neelakantan, Shyam, Sastry, Askell, Agarwal, Herbert-Voss, Krueger, Henighan, Child, Ramesh, Ziegler, Wu, Winter, Hesse, Chen, Sigler, Litwin, Gray, Chess, Clark, Berner, McCandlish, Radford, Sutskever, and Amodei}]{brown-etal-2020-language}
Tom Brown, Benjamin Mann, Nick Ryder, Melanie Subbiah, Jared~D Kaplan, Prafulla Dhariwal, Arvind Neelakantan, Pranav Shyam, Girish Sastry, Amanda Askell, Sandhini Agarwal, Ariel Herbert-Voss, Gretchen Krueger, Tom Henighan, Rewon Child, Aditya Ramesh, Daniel Ziegler, Jeffrey Wu, Clemens Winter, and 12 others. 2020.
\newblock \href {https://proceedings.neurips.cc/paper_files/paper/2020/file/1457c0d6bfcb4967418bfb8ac142f64a-Paper.pdf} {Language models are few-shot learners}.
\newblock In \emph{Advances in Neural Information Processing Systems}, volume~33, pages 1877--1901. Curran Associates, Inc.

\bibitem[{Bulte and Tezcan(2019)}]{bulte-tezcan-2019-neural}
Bram Bulte and Arda Tezcan. 2019.
\newblock \href {https://doi.org/10.18653/v1/P19-1175} {Neural fuzzy repair: Integrating fuzzy matches into neural machine translation}.
\newblock In \emph{Proceedings of the 57th Annual Meeting of the Association for Computational Linguistics}, pages 1800--1809, Florence, Italy. Association for Computational Linguistics.

\bibitem[{Burlot and Yvon(2018)}]{burlot-yvon-2018-using}
Franck Burlot and Fran{\c{c}}ois Yvon. 2018.
\newblock \href {https://doi.org/10.18653/v1/W18-6315} {Using monolingual data in neural machine translation: a systematic study}.
\newblock In \emph{Proceedings of the Third Conference on Machine Translation: Research Papers}, pages 144--155, Brussels, Belgium. Association for Computational Linguistics.

\bibitem[{Cai et~al.(2021)Cai, Wang, Li, Lam, and Liu}]{cai-etal-2021-neural}
Deng Cai, Yan Wang, Huayang Li, Wai Lam, and Lemao Liu. 2021.
\newblock \href {https://doi.org/10.18653/v1/2021.acl-long.567} {Neural machine translation with monolingual translation memory}.
\newblock In \emph{Proceedings of the 59th Annual Meeting of the Association for Computational Linguistics and the 11th International Joint Conference on Natural Language Processing (Volume 1: Long Papers)}, pages 7307--7318, Online. Association for Computational Linguistics.

\bibitem[{Cakir et~al.(2019)Cakir, He, Xia, Kulis, and Sclaroff}]{cakir-etal-2019-deep}
Fatih Cakir, Kun He, Xide Xia, Brian Kulis, and Stan Sclaroff. 2019.
\newblock \href {https://openaccess.thecvf.com/content_CVPR_2019/papers/Cakir_Deep_Metric_Learning_to_Rank_CVPR_2019_paper.pdf} {Deep metric learning to rank}.
\newblock \emph{Proceedings of the IEEE/CVF Conference on Computer Vision and Pattern Recognition (CVPR)}.

\bibitem[{Cao et~al.(2007)Cao, Qin, Liu, Tsai, and Li}]{cao-etal-2016-learning}
Zhe Cao, Tao Qin, Tie-Yan Liu, Ming-Feng Tsai, and Hang Li. 2007.
\newblock \href {https://doi.org/10.1145/1273496.1273513} {Learning to rank: from pairwise approach to listwise approach}.
\newblock In \emph{Proceedings of the 24th International Conference on Machine Learning}, ICML '07, page 129–136, New York, NY, USA. Association for Computing Machinery.

\bibitem[{Caswell et~al.(2019)Caswell, Chelba, and Grangier}]{caswell-etal-2019-tagged}
Isaac Caswell, Ciprian Chelba, and David Grangier. 2019.
\newblock \href {https://doi.org/10.18653/v1/W19-5206} {Tagged back-translation}.
\newblock In \emph{Proceedings of the Fourth Conference on Machine Translation (Volume 1: Research Papers)}, pages 53--63, Florence, Italy. Association for Computational Linguistics.

\bibitem[{Cheng et~al.(2022)Cheng, Gao, Liu, Zhao, and Yan}]{cheng-etal-2022-neural}
Xin Cheng, Shen Gao, Lemao Liu, Dongyan Zhao, and Rui Yan. 2022.
\newblock \href {https://doi.org/10.18653/v1/2022.emnlp-main.235} {Neural machine translation with contrastive translation memories}.
\newblock In \emph{Proceedings of the 2022 Conference on Empirical Methods in Natural Language Processing}, pages 3591--3601, Abu Dhabi, United Arab Emirates. Association for Computational Linguistics.

\bibitem[{Costa-jussà et~al.(2024)Costa-jussà, Cross, Çelebi, Elbayad, Heafield, Heffernan, Kalbassi, Lam, Licht, Maillard, Sun, Wang, Wenzek, Youngblood, Akula, Barrault, Gonzalez, Hansanti, Hoffman, Jarrett, Sadagopan, Rowe, Spruit, Tran, Andrews, Ayan, Bhosale, Edunov, Fan, Gao, Goswami, Guzmán, Koehn, Mourachko, Ropers, Saleem, Schwenk, Wang, and Team}]{costa-etal-2024-scaling}
Marta~R. Costa-jussà, James Cross, Onur Çelebi, Maha Elbayad, Kenneth Heafield, Kevin Heffernan, Elahe Kalbassi, Janice Lam, Daniel Licht, Jean Maillard, Anna Sun, Skyler Wang, Guillaume Wenzek, Al~Youngblood, Bapi Akula, Loic Barrault, Gabriel~Mejia Gonzalez, Prangthip Hansanti, John Hoffman, and 20 others. 2024.
\newblock \href {https://doi.org/10.1038/s41586-024-07335-x} {Scaling neural machine translation to 200 languages}.
\newblock \emph{Nature}, 630(8018):841--846.
\newblock ISBN: 1476-4687 tex.date-added: 2024-08-21 15:32:57 +0200 tex.date-modified: 2024-08-21 15:32:57 +0200.

\bibitem[{Currey et~al.(2017)Currey, Miceli~Barone, and Heafield}]{currey-etal-2017-copied}
Anna Currey, Antonio~Valerio Miceli~Barone, and Kenneth Heafield. 2017.
\newblock \href {https://doi.org/10.18653/v1/W17-4715} {Copied monolingual data improves low-resource neural machine translation}.
\newblock In \emph{Proceedings of the Second Conference on Machine Translation}, pages 148--156, Copenhagen, Denmark. Association for Computational Linguistics.

\bibitem[{Dettmers et~al.(2023)Dettmers, Pagnoni, Holtzman, and Zettlemoyer}]{dettmers-etal-2023-qlora}
Tim Dettmers, Artidoro Pagnoni, Ari Holtzman, and Luke Zettlemoyer. 2023.
\newblock \href {https://arxiv.org/abs/2305.14314} {Qlora: Efficient finetuning of quantized llms}.
\newblock \emph{CoRR}, abs/2305.14314.

\bibitem[{Devlin et~al.(2019)Devlin, Chang, Lee, and Toutanova}]{devlin-etal-2019-bert}
Jacob Devlin, Ming-Wei Chang, Kenton Lee, and Kristina Toutanova. 2019.
\newblock \href {https://doi.org/10.18653/v1/N19-1423} {{BERT}: Pre-training of deep bidirectional transformers for language understanding}.
\newblock In \emph{Proceedings of the 2019 Conference of the North {A}merican Chapter of the Association for Computational Linguistics: Human Language Technologies, Volume 1 (Long and Short Papers)}, pages 4171--4186, Minneapolis, Minnesota. Association for Computational Linguistics.

\bibitem[{Dorr(1994)}]{dorr-1994-machine}
Bonnie~J. Dorr. 1994.
\newblock \href {https://aclanthology.org/J94-4004/} {Machine translation divergences: A formal description and proposed solution}.
\newblock \emph{Computational Linguistics}, 20(4):597--633.

\bibitem[{Douze et~al.(2024)Douze, Guzhva, Deng, Johnson, Szilvasy, Mazaré, Lomeli, Hosseini, and Jégou}]{douze-etal-2024-faiss}
Matthijs Douze, Alexandr Guzhva, Chengqi Deng, Jeff Johnson, Gergely Szilvasy, Pierre-Emmanuel Mazaré, Maria Lomeli, Lucas Hosseini, and Hervé Jégou. 2024.
\newblock \href {https://arxiv.org/abs/2401.08281} {The {Faiss} library}.
\newblock \emph{CoRR}, abs/2401.08281.

\bibitem[{Duquenne et~al.(2023)Duquenne, Schwenk, and Sagot}]{ambroise-etal-2023-sonar}
Paul-Ambroise Duquenne, Holger Schwenk, and Benoît Sagot. 2023.
\newblock \href {https://arxiv.org/abs/2308.11466} {Sonar: Sentence-level multimodal and language-agnostic representations}.
\newblock \emph{CoRR}, abs/2308.11466.

\bibitem[{Edunov et~al.(2018)Edunov, Ott, Auli, and Grangier}]{edunov-etal-2018-understanding}
Sergey Edunov, Myle Ott, Michael Auli, and David Grangier. 2018.
\newblock \href {https://doi.org/10.18653/v1/D18-1045} {Understanding back-translation at scale}.
\newblock In \emph{Proceedings of the 2018 Conference on Empirical Methods in Natural Language Processing}, pages 489--500, Brussels, Belgium. Association for Computational Linguistics.

\bibitem[{Feng et~al.(2022)Feng, Yang, Cer, Arivazhagan, and Wang}]{feng-etal-2022-language}
Fangxiaoyu Feng, Yinfei Yang, Daniel Cer, Naveen Arivazhagan, and Wei Wang. 2022.
\newblock \href {https://doi.org/10.18653/v1/2022.acl-long.62} {Language-agnostic {BERT} sentence embedding}.
\newblock In \emph{Proceedings of the 60th Annual Meeting of the Association for Computational Linguistics (Volume 1: Long Papers)}, pages 878--891, Dublin, Ireland. Association for Computational Linguistics.

\bibitem[{Gillick et~al.(2018)Gillick, Presta, and Tomar}]{gillick-etal-2018-end}
Daniel Gillick, Alessandro Presta, and Gaurav~Singh Tomar. 2018.
\newblock \href {https://arxiv.org/abs/1811.08008} {End-to-end retrieval in continuous space}.
\newblock \emph{CoRR}, abs/1811.08008.

\bibitem[{Gu et~al.(2019)Gu, Wang, and Zhao}]{gu-etal-2019-levenshtein}
Jiatao Gu, Changhan Wang, and Junbo Zhao. 2019.
\newblock \href {https://proceedings.neurips.cc/paper/2019/file/675f9820626f5bc0afb47b57890b466e-Paper.pdf} {Levenshtein transformer}.
\newblock In \emph{Advances in Neural Information Processing Systems}, volume~32. Curran Associates, Inc.

\bibitem[{Gu et~al.(2018)Gu, Wang, Cho, and Li}]{gu-etal-2018-search}
Jiatao Gu, Yong Wang, Kyunghyun Cho, and Victor~O.K. Li. 2018.
\newblock \href {https://doi.org/10.1609/aaai.v32i1.12013} {Search {Engine} {Guided} {Neural} {Machine} {Translation}}.
\newblock \emph{Proceedings of the AAAI Conference on Artificial Intelligence}, 32(1).

\bibitem[{He et~al.(2021)He, Huang, Cui, Li, and Liu}]{he-etal-2021-fast}
Qiuxiang He, Guoping Huang, Qu~Cui, Li~Li, and Lemao Liu. 2021.
\newblock \href {https://doi.org/10.18653/v1/2021.acl-long.246} {Fast and accurate neural machine translation with translation memory}.
\newblock In \emph{Proceedings of the 59th Annual Meeting of the Association for Computational Linguistics and the 11th International Joint Conference on Natural Language Processing (Volume 1: Long Papers)}, pages 3170--3180, Online. Association for Computational Linguistics.

\bibitem[{Hendy et~al.(2023)Hendy, Abdelrehim, Sharaf, Raunak, Gabr, Matsushita, Kim, Afify, and Awadalla}]{hendy-etal-2023-howgood}
Amr Hendy, Mohamed Abdelrehim, Amr Sharaf, Vikas Raunak, Mohamed Gabr, Hitokazu Matsushita, Young~Jin Kim, Mohamed Afify, and Hany~Hassan Awadalla. 2023.
\newblock \href {https://doi.org/10.48550/ARXIV.2302.09210} {How good are {GPT} models at machine translation? a comprehensive evaluation}.
\newblock \emph{CoRR}, abs/2302.09210.

\bibitem[{Kay(1997)}]{kay-1997-proper}
Martin Kay. 1997.
\newblock \href {http://www.jstor.org/stable/40009025} {The proper place of men and machines in language translation}.
\newblock \emph{Machine Translation}, 12(1/2):3--23.

\bibitem[{Koehn and Senellart(2010)}]{koehn-senellart-2010-convergence}
Philipp Koehn and Jean Senellart. 2010.
\newblock \href {https://aclanthology.org/2010.jec-1.4/} {Convergence of translation memory and statistical machine translation}.
\newblock In \emph{Proceedings of the Second Joint EM+/CNGL Workshop: Bringing MT to the User: Research on Integrating MT in the Translation Industry}, pages 21--32, Denver, Colorado, USA. Association for Machine Translation in the Americas.

\bibitem[{Kulis(2013)}]{kulis-2013-metric}
Brian Kulis. 2013.
\newblock \href {https://doi.org/10.1561/2200000019} {Metric learning: A survey}.
\newblock \emph{Foundations and Trends® in Machine Learning}, 5(4):287--364.

\bibitem[{Lample et~al.(2018)Lample, Ott, Conneau, Denoyer, and Ranzato}]{lample-etal-2018-phrase}
Guillaume Lample, Myle Ott, Alexis Conneau, Ludovic Denoyer, and Marc{'}Aurelio Ranzato. 2018.
\newblock \href {https://doi.org/10.18653/v1/D18-1549} {Phrase-based {\&} neural unsupervised machine translation}.
\newblock In \emph{Proceedings of the 2018 Conference on Empirical Methods in Natural Language Processing}, pages 5039--5049, Brussels, Belgium. Association for Computational Linguistics.

\bibitem[{Levenshtein(1965)}]{levenshtein-1965-binary}
Vladimir~I. Levenshtein. 1965.
\newblock Binary codes capable of correcting deletions, insertions, and reversals.
\newblock \emph{Soviet physics. Doklady}, 10:707--710.

\bibitem[{Li et~al.(2022)Li, Su, Cai, Wang, and Liu}]{li-etal-2023-survey}
Huayang Li, Yixuan Su, Deng Cai, Yan Wang, and Lemao Liu. 2022.
\newblock \href {https://arxiv.org/abs/2202.01110} {A survey on retrieval-augmented text generation}.
\newblock \emph{CoRR}, abs/2202.01110.

\bibitem[{Martins et~al.(2025)Martins, Alves, Fernandes, Guerreiro, Rei, Farajian, Klimaszewski, Alves, Pombal, Boizard, Faysse, Colombo, Yvon, Haddow, de~Souza, Birch, and Martins}]{martins-etal-2025-eurollm9b}
Pedro~Henrique Martins, João Alves, Patrick Fernandes, Nuno~M. Guerreiro, Ricardo Rei, Amin Farajian, Mateusz Klimaszewski, Duarte~M. Alves, José Pombal, Nicolas Boizard, Manuel Faysse, Pierre Colombo, François Yvon, Barry Haddow, José G.~C. de~Souza, Alexandra Birch, and André F.~T. Martins. 2025.
\newblock \href {https://arxiv.org/abs/2506.04079} {Eurollm-9b: Technical report}.
\newblock \emph{Preprint}, arXiv:2506.04079.

\bibitem[{Moslem et~al.(2023)Moslem, Haque, Kelleher, and Way}]{moslem-etal-2023-adaptive}
Yasmin Moslem, Rejwanul Haque, John~D. Kelleher, and Andy Way. 2023.
\newblock \href {https://aclanthology.org/2023.eamt-1.22/} {Adaptive machine translation with large language models}.
\newblock In \emph{Proceedings of the 24th Annual Conference of the European Association for Machine Translation}, pages 227--237, Tampere, Finland. European Association for Machine Translation.

\bibitem[{Niwa et~al.(2022)Niwa, Takase, and Okazaki}]{niwa-etal-2022-nearest}
Ayana Niwa, Sho Takase, and Naoaki Okazaki. 2022.
\newblock \href {https://doi.org/10.48550/ARXIV.2208.12496} {Nearest neighbor non-autoregressive text generation}.
\newblock \emph{CoRR}, abs/2208.12496.

\bibitem[{Ott et~al.(2019)Ott, Edunov, Baevski, Fan, Gross, Ng, Grangier, and Auli}]{ott-etal-2019-fairseq}
Myle Ott, Sergey Edunov, Alexei Baevski, Angela Fan, Sam Gross, Nathan Ng, David Grangier, and Michael Auli. 2019.
\newblock \href {https://doi.org/10.18653/v1/N19-4009} {fairseq: A fast, extensible toolkit for sequence modeling}.
\newblock In \emph{Proceedings of the 2019 Conference of the North {A}merican Chapter of the Association for Computational Linguistics (Demonstrations)}, pages 48--53, Minneapolis, Minnesota. Association for Computational Linguistics.

\bibitem[{Papineni et~al.(2002)Papineni, Roukos, Ward, and Zhu}]{papineni-etal-2002-bleu}
Kishore Papineni, Salim Roukos, Todd Ward, and Wei-Jing Zhu. 2002.
\newblock \href {https://doi.org/10.3115/1073083.1073135} {{B}leu: a method for automatic evaluation of machine translation}.
\newblock In \emph{Proceedings of the 40th Annual Meeting of the Association for Computational Linguistics}, pages 311--318, Philadelphia, Pennsylvania, USA. Association for Computational Linguistics.

\bibitem[{Pham et~al.(2020)Pham, Xu, Crego, Yvon, and Senellart}]{pham-etal-2020-priming}
Minh~Quang Pham, Jitao Xu, Josep Crego, Fran{\c{c}}ois Yvon, and Jean Senellart. 2020.
\newblock \href {https://aclanthology.org/2020.wmt-1.63/} {Priming neural machine translation}.
\newblock In \emph{Proceedings of the Fifth Conference on Machine Translation}, pages 516--527, Online. Association for Computational Linguistics.

\bibitem[{Post(2018)}]{post-2018-call}
Matt Post. 2018.
\newblock \href {https://doi.org/10.18653/v1/W18-6319} {A call for clarity in reporting {BLEU} scores}.
\newblock In \emph{Proceedings of the Third Conference on Machine Translation: Research Papers}, pages 186--191, Brussels, Belgium. Association for Computational Linguistics.

\bibitem[{Radford et~al.(2019)Radford, Wu, Child, Luan, Amodei, Sutskever et~al.}]{radford-etal-2019-language}
Alec Radford, Jeffrey Wu, Rewon Child, David Luan, Dario Amodei, Ilya Sutskever, and 1 others. 2019.
\newblock Language models are unsupervised multitask learners.
\newblock \emph{OpenAI blog}, 1(8):9.

\bibitem[{Reheman et~al.(2023)Reheman, Zhou, Luo, Yang, Xiao, and Zhu}]{reheman-etal-2023-prompting}
Abudurexiti Reheman, Tao Zhou, Yingfeng Luo, Di~Yang, Tong Xiao, and Jingbo Zhu. 2023.
\newblock \href {https://doi.org/10.1609/aaai.v37i11.26585} {Prompting neural machine translation with translation memories}.
\newblock \emph{Proceedings of the AAAI Conference on Artificial Intelligence}, 37(11):13519--13527.

\bibitem[{Rei et~al.(2020)Rei, Stewart, Farinha, and Lavie}]{rei-etal-2020-comet}
Ricardo Rei, Craig Stewart, Ana~C Farinha, and Alon Lavie. 2020.
\newblock \href {https://doi.org/10.18653/v1/2020.emnlp-main.213} {{COMET}: A neural framework for {MT} evaluation}.
\newblock In \emph{Proceedings of the 2020 Conference on Empirical Methods in Natural Language Processing (EMNLP)}, pages 2685--2702, Online. Association for Computational Linguistics.

\bibitem[{Rei et~al.(2022)Rei, Treviso, Guerreiro, Zerva, Farinha, Maroti, C.~de Souza, Glushkova, Alves, Coheur, Lavie, and Martins}]{rei-etal-2022-cometkiwi}
Ricardo Rei, Marcos Treviso, Nuno~M. Guerreiro, Chrysoula Zerva, Ana~C Farinha, Christine Maroti, Jos{\'e}~G. C.~de Souza, Taisiya Glushkova, Duarte Alves, Luisa Coheur, Alon Lavie, and Andr{\'e} F.~T. Martins. 2022.
\newblock \href {https://aclanthology.org/2022.wmt-1.60/} {{C}omet{K}iwi: {IST}-unbabel 2022 submission for the quality estimation shared task}.
\newblock In \emph{Proceedings of the Seventh Conference on Machine Translation (WMT)}, pages 634--645, Abu Dhabi, United Arab Emirates (Hybrid). Association for Computational Linguistics.

\bibitem[{Robertson and Walker(1994)}]{robertson-walker-1994-some}
Stephen Robertson and Steve Walker. 1994.
\newblock \href {https://doi.org/10.1007/978-1-4471-2099-5_24} {Some simple effective approximations to the {2-Poisson} model for probabilistic weighted retrieval}.
\newblock In \emph{Proceedings of the 17th ACM Conference on Research and Development in Information Retrieval (SIGIR), Dublin, Ireland}, pages 232--241.

\bibitem[{Rudin(2019)}]{rudin-cynthia-2019-stop}
Cynthia Rudin. 2019.
\newblock \href {https://doi.org/10.1038/s42256-019-0048-x} {Stop explaining black box machine learning models for high stakes decisions and use interpretable models instead}.
\newblock \emph{Nature Machine Intelligence}, 1(5):206--215.

\bibitem[{Schultz and Joachims(2003)}]{schultz-etal-2003-advances}
Matthew Schultz and Thorsten Joachims. 2003.
\newblock \href {https://proceedings.neurips.cc/paper_files/paper/2003/file/d3b1fb02964aa64e257f9f26a31f72cf-Paper.pdf} {Learning a distance metric from relative comparisons}.
\newblock In \emph{Advances in Neural Information Processing Systems}, volume~16. MIT Press.

\bibitem[{Sennrich et~al.(2016{\natexlab{a}})Sennrich, Haddow, and Birch}]{sennrich-etal-2016-improving}
Rico Sennrich, Barry Haddow, and Alexandra Birch. 2016{\natexlab{a}}.
\newblock \href {https://doi.org/10.18653/v1/P16-1009} {Improving neural machine translation models with monolingual data}.
\newblock In \emph{Proceedings of the 54th Annual Meeting of the Association for Computational Linguistics (Volume 1: Long Papers)}, pages 86--96, Berlin, Germany. Association for Computational Linguistics.

\bibitem[{Sennrich et~al.(2016{\natexlab{b}})Sennrich, Haddow, and Birch}]{sennrich-etal-2016-neural}
Rico Sennrich, Barry Haddow, and Alexandra Birch. 2016{\natexlab{b}}.
\newblock \href {https://doi.org/10.18653/v1/P16-1162} {Neural machine translation of rare words with subword units}.
\newblock In \emph{Proceedings of the 54th Annual Meeting of the Association for Computational Linguistics (Volume 1: Long Papers)}, pages 1715--1725, Berlin, Germany. Association for Computational Linguistics.

\bibitem[{Sohn(2016)}]{sohn-2016-improved}
Kihyuk Sohn. 2016.
\newblock \href {https://proceedings.neurips.cc/paper_files/paper/2016/file/6b180037abbebea991d8b1232f8a8ca9-Paper.pdf} {Improved deep metric learning with multi-class n-pair loss objective}.
\newblock In \emph{Advances in Neural Information Processing Systems}, volume~29. Curran Associates, Inc.

\bibitem[{Suárez et~al.(2021)Suárez, García, and Herrera}]{suarez-etal-2021-atutorial}
Juan~Luis Suárez, Salvador García, and Francisco Herrera. 2021.
\newblock \href {https://doi.org/10.1016/j.neucom.2020.08.017} {A tutorial on distance metric learning: Mathematical foundations, algorithms, experimental analysis, prospects and challenges}.
\newblock \emph{Neurocomputing}, 425:300--322.

\bibitem[{Tamura et~al.(2023)Tamura, Wang, Utsuro, and Nagata}]{tamura-etal-2023-target}
Takuya Tamura, Xiaotian Wang, Takehito Utsuro, and Masaaki Nagata. 2023.
\newblock \href {https://aclanthology.org/2023.mtsummit-research.26/} {Target language monolingual translation memory based {NMT} by cross-lingual retrieval of similar translations and reranking}.
\newblock In \emph{Proceedings of Machine Translation Summit XIX, Vol. 1: Research Track}, pages 313--323, Macau SAR, China. Asia-Pacific Association for Machine Translation.

\bibitem[{Tezcan et~al.(2024)Tezcan, Skidanova, and Moerman}]{tezcan-etal-2024-improving}
Arda Tezcan, Alina Skidanova, and Thomas Moerman. 2024.
\newblock \href {https://doi.org/10.14712/00326585.030} {Improving fuzzy match augmented neural machine translation in specialised domains through synthetic data}.
\newblock \emph{The Prague Bulletin of Mathematical Linguistics}, 122:9--42.

\bibitem[{Tiedemann(2012)}]{tiedemann-2012-parallel}
J{\"o}rg Tiedemann. 2012.
\newblock \href {https://aclanthology.org/L12-1246/} {Parallel data, tools and interfaces in {OPUS}}.
\newblock In \emph{Proceedings of the Eighth International Conference on Language Resources and Evaluation ({LREC}`12)}, pages 2214--2218, Istanbul, Turkey. European Language Resources Association (ELRA).

\bibitem[{Vaswani et~al.(2017)Vaswani, Shazeer, Parmar, Uszkoreit, Jones, Gomez, Kaiser, and Polosukhin}]{vaswani-etal-2017-attention}
Ashish Vaswani, Noam Shazeer, Niki Parmar, Jakob Uszkoreit, Llion Jones, Aidan~N. Gomez, \L{}ukasz Kaiser, and Illia Polosukhin. 2017.
\newblock \href {https://papers.nips.cc/paper_files/paper/2017/hash/3f5ee243547dee91fbd053c1c4a845aa-Abstract.html} {Attention is all you need}.
\newblock In \emph{Proceedings of the 31st International Conference on Neural Information Processing Systems}, NIPS'17, page 6000–6010, Red Hook, NY, USA. Curran Associates Inc.

\bibitem[{Vilar et~al.(2023)Vilar, Freitag, Cherry, Luo, Ratnakar, and Foster}]{vilar-etal-2023-prompting}
David Vilar, Markus Freitag, Colin Cherry, Jiaming Luo, Viresh Ratnakar, and George Foster. 2023.
\newblock \href {https://doi.org/10.18653/v1/2023.acl-long.859} {Prompting {P}a{LM} for translation: Assessing strategies and performance}.
\newblock In \emph{Proceedings of the 61st Annual Meeting of the Association for Computational Linguistics (Volume 1: Long Papers)}, pages 15406--15427, Toronto, Canada. Association for Computational Linguistics.

\bibitem[{Wang et~al.(2013)Wang, Wang, Li, He, and Liu}]{wang-etal-2013-theoretical}
Yining Wang, Liwei Wang, Yuanzhi Li, Di~He, and Tie{-}Yan Liu. 2013.
\newblock \href {https://proceedings.mlr.press/v30/Wang13.html} {A theoretical analysis of {NDCG} type ranking measures}.
\newblock In \emph{Proceedings of the 26th Annual Conference on Learning Theory}, volume~30 of \emph{Proceedings of Machine Learning Research}, pages 25--54, Princeton, NJ, USA. PMLR.

\bibitem[{Xia et~al.(2019)Xia, Huang, Liu, and Shi}]{xia-etal-2019-graph}
Mengzhou Xia, Guoping Huang, Lemao Liu, and Shuming Shi. 2019.
\newblock \href {https://doi.org/10.1609/aaai.v33i01.33017297} {Graph based translation memory for neural machine translation}.
\newblock \emph{Proceedings of the AAAI Conference on Artificial Intelligence}, 33(01):7297--7304.

\bibitem[{Xu et~al.(2020)Xu, Crego, and Senellart}]{xu-etal-2020-boosting}
Jitao Xu, Josep Crego, and Jean Senellart. 2020.
\newblock \href {https://doi.org/10.18653/v1/2020.acl-main.144} {Boosting neural machine translation with similar translations}.
\newblock In \emph{Proceedings of the 58th Annual Meeting of the Association for Computational Linguistics}, pages 1580--1590, Online. Association for Computational Linguistics.

\bibitem[{Xu et~al.(2023)Xu, Crego, and Yvon}]{xu-etal-2023-integrating}
Jitao Xu, Josep Crego, and Fran{\c{c}}ois Yvon. 2023.
\newblock \href {https://doi.org/10.18653/v1/2023.eacl-main.96} {Integrating translation memories into non-autoregressive machine translation}.
\newblock In \emph{Proceedings of the 17th Conference of the European Chapter of the Association for Computational Linguistics}, pages 1326--1338, Dubrovnik, Croatia. Association for Computational Linguistics.

\bibitem[{Zebaze et~al.(2024)Zebaze, Sagot, and Bawden}]{zebaze-etal-2024-context}
Armel Zebaze, Benoît Sagot, and Rachel Bawden. 2024.
\newblock \href {https://arxiv.org/abs/2408.00397} {In-context example selection via similarity search improves low-resource machine translation}.
\newblock \emph{CoRR}, abs/2408.00397.

\bibitem[{Zhang et~al.(2023)Zhang, Haddow, and Birch}]{zhang-et-al-2023-prompting}
Biao Zhang, Barry Haddow, and Alexandra Birch. 2023.
\newblock Prompting large language model for machine translation: A case study.
\newblock In \emph{Proceedings of the 40th International Conference on Machine Learning}, ICML'23. JMLR.org.

\bibitem[{Zheng et~al.(2023)Zheng, Wang, Wang, Chen, Zhang, and Tu}]{zheng-etal-2023-towards}
Kangjie Zheng, Longyue Wang, Zhihao Wang, Binqi Chen, Ming Zhang, and Zhaopeng Tu. 2023.
\newblock \href {https://doi.org/10.1109/ICASSP49357.2023.10094646} {Towards a unified training for {Levenshtein} transformer}.
\newblock In \emph{Proceedings of the IEEE International Conference on Acoustics, Speech and Signal Processing (ICASSP)}, pages 1--5.

\end{thebibliography}
% \bibliography{local}

\appendix

\section{Parallel Data Specification} \label{appendix:data}

\begin{table*}[ht]
	\centering
	\footnotesize
	\resizebox{\textwidth}{!}{
		\begin{tabular}{lrrrrrrrrrrrrrrrrr}
			\toprule
			     & \multicolumn{11}{c}{English-French} & \multicolumn{5}{c}{German-English}                                                                                               \\
			\cmidrule(lr){2-12}\cmidrule(lr){13-17}
			     & ECB                                 & EME                                & Epp  & GNO & JRC  & KDE  & News & PHP & TED  & Ubu & Wiki & KDE  & Kor & JRC  & EME  & Sub  \\
			\cmidrule(lr){2-12}\cmidrule(lr){13-17}
			size & 149k                                & 272k                               & 1,9M & 44k & 492k & 126k & 133k & 7k  & 148k & 6k  & 597k & 223k & 18k & 467k & 248k & 500k \\
			\%   & 3.8                                 & 7.0                                & 49.0 & 1.1 & 12.7 & 3.3  & 3.4  & 0.2 & 3.8  & 0.2 & 15.4 & 15,3 & 1,2 & 32,1 & 17,0 & 34,3 \\
			\bottomrule
			\normalsize
		\end{tabular}
	}
	\caption{\label{tab:data-stats} Size of each domain, and its proportion w.r.t. the size of the corpus of the same language pair.}
\end{table*}

The en-fr data is a clean version\footnote{We filter noisy parallel pairs with COMETKiwi \cite{rei-etal-2022-cometkiwi} and prepare a new train/valid/test split.} of the one used by \citep{xu-etal-2020-boosting}. Validation and test sets each contain $1,000$ segments.
% Since French is morphologically more complex than English, there is a high motivation to directly retrieve in the target language segments where words are already correctly inflected.
The de-en corpus is borrowed unchanged from \citep{aharoni-goldberg-2020-unsupervised} and is also used in \citep{cai-etal-2021-neural,agrawal-etal-2023-context,bouthors-etal-2024-retrieving}. The associated validation and test sets contain $2,000$ segments. The per-domain corpus sizes are available in table~\ref{tab:data-stats}.
The en-uk corpus is the one provided by \citep{tezcan-etal-2024-improving}, containing about 286K training sentences, 2000 validation sentences and 1898 test sentences. It corresponds to various legal texts.

Note that the preprocessing stages were performed with \textit{subword-nmt} \citep{sennrich-etal-2016-neural} for tokenization and BPE, and \textit{fairseq} \citep{ott-etal-2019-fairseq} for binarization.

\section{Search Parameterization} \label{appendix:search}

For both methods, we search up to $k$=3 examples.

\subsection{Fuzzy Matching}

The execution of the fuzzy matching methods (\fuzzysrc, \fuzzybt and \fuzzygold) is performed using an open source library\footnote{https://github.com/SYSTRAN/fuzzy-match}.
During the evaluation, there is no filter applied before scoring the segments with \Lev. However, the examples associated to the training samples are selected with \Lev and a BM25 filter. This ensures a reasonable time to build the set of examples for training.

\subsection{FAISS}

The execution of the CLIR techniques is done with library FAISS \citep{douze-etal-2024-faiss}.
The $k$NN search is performed on GPU (IndexFlatIP), without IVF (for faster search) or quantizer (for lower memory footprint) to ensure the retrieval of the best matches. Indeed, both optimizations can prevent the model from finding the optimal $k$ nearest neighbors.

\section{NMT Architectures Configuration}

Both \mlevt{3} and \NFA models rely on the default backbone Transformer architecture \citep{vaswani-etal-2017-attention}. We train one instance of \mlevt{3} for each language pair (en-fr and de-en), using only the training data available in the corresponding datasets. Our \NFA models are trained on much larger sets of parallel data, including a good share of Web data. We use these models to showcase the effectiveness of cross-lingual retrieval for RAMNT systems trained with large, high-quality, parallel datasets (8M sentences for en-fr and 11.5M for de-en). After inspecting the training datasets of \NFA models, we found that some of them partly overlap with the training data; the corresponding contamination rates are reported in Table~\ref{tab:bleu-per-domain-nfa-clir}.

As we compare retrieval techniques \emph{for fixed MT architectures}, these overlaps do not affect on our main conclusions (see also the discussion in \textsection{\ref{par:data-contamination}}).

% Note that there is a case of contamination of sole of the sentences in the test sets of some domains. Each sentence has a probability to have been sampled for the training set of \NFA, corresponding to a

In addition, we use the multilingual NLLB model \citep{costa-etal-2024-scaling} out-of-the-box for both language pairs.\footnote{version: \texttt{NLLB-200-distilled-1.3B}\\\url{https://huggingface.co/facebook/nllb-200-1.3B}} This large encoder-decoder model is used to generate the back-translation data; it also provides a baseline against which to appreciate the performance of RANMT architectures.
% , it is a custom model trained on a large amount of parallel data that obtains higher scores than one trained as \mlevt{3}.
As for the \NFA models, the training data of NLLB partly overlaps with our test sets, for domains ECB, EMEA, JRC-Acquis and OpenSubtitles. Contamination is discussed in \textsection{\ref{par:data-contamination}}.

\section{Adapting \EuroLLM} \label{appendix:eurollm}

LLMs are well suited for in-context learning in $k$-shot settings \citep{radford-etal-2019-language}: a set of $k$ demonstrations of the task are presented to the model so that the completion of the task fulfills the prompt pattern/template. In our case, the situation is slightly different as we \emph{only include target-side examples}, which are not suitable demonstrations of the translation task.

To adapt an LLM to this setting, we first compare (a) prompting and (b) fine-tuning strategies on the Wikipedia domain.\footnote{Another alternative, that we do not consider in this work, would be to use backtranslation to automatically craft the missing source for in-context examples.}

Regarding (a), we try to take advantage of the in-context abilities of the \emph{Instruct} version of \EuroLLM-9B, simulating in-context examples with the follow prompt patterns:
\begin{itemize}
	\item \textbf{naive}:\\
	      \textit{Consider the following \textbf{French} similar translations.} \\
	      \textit{\{k examples in \textbf{French}\}} \\
	      \textit{Translate the following text from \textbf{English} to \textbf{French}.} \\
	      \textit{\textbf{English}:} \\
	      \textit{\{source text\}} \\
	      \textit{\textbf{French}:}
	\item \textbf{empty:}\\
	      \textit{Consider the following \textbf{English}-\textbf{French} examples.} \\
	      \textit{\textbf{English}:} \\
	      \\
	      \textit{\textbf{French}:} \\
	      \textit{\{\textbf{French} example \#1\}} \\
	      \textit{[...]} \\
	      \textit{Translate the following text from \textbf{English} to \textbf{French}.} \\
	      \textit{\textbf{English}:} \\
	      \textit{\{source\}} \\
	      \textit{\textbf{French}:}
	\item \textbf{copy:}\\
	      \textit{Consider the following \textbf{English}-\textbf{French} examples.} \\
	      \textit{\textbf{English}:} \\
	      \textit{\{\textbf{French} example \#1\}} \\
	      \textit{\textbf{French}:} \\
	      \textit{\{\textbf{French} example \#1\}} \\
	      \textit{[...]} \\
	      \textit{Translate the following text from \textbf{English} to \textbf{French}.} \\
	      \textit{\textbf{English}:} \\
	      \textit{\{source\}} \\
	      \textit{\textbf{French}:}
\end{itemize}
The \textbf{naive} prompt gives a simple description of the task, where only the target side of examples are introduced. The \textbf{copy} and \textbf{empty} prompts simulate a standard $k$-shot prompt, where the missing source text is either left empty or contains a copy of the target text. We compare these prompts with a \textbf{0-shot} setting, where no example is provided to the model (starting directly with "\textit{Translate the following text}" for any of the three templates).

Regarding (b), we fine-tune the \emph{base} \EuroLLM-9B model with the \textbf{naive} prompt with up to 20K sentences from the training set for each domain. Fine-tuning is performed with QLoRA \citep{dettmers-etal-2023-qlora} with a $r$=16, $\alpha$=32, dropout=0.05, no bias, applied on all linear layers, and 4-bit quantization. Training is performed with huggingface implementation\footnote{https://huggingface.co/docs/transformers/trainer} with a batch size of 32 for a single epoch and a learning rate of 2e-10. The in-context examples used for training are retrieved using the \fuzzysrc strategy.

\begin{table}[!h]
	\centering
	\begin{tabular}{ccccc}
		\toprule
		  \multicolumn{4}{c}{Instruct} & \multicolumn{1}{c}{\textbf{ft}-base}                          \\
		\cmidrule(lr){1-4}\cmidrule(lr){5-5}
          0-shot & naive & copy & empty & naive  \\
          (k=0) & (k=3) & (k=3) & (k=3) & (k=3) \\
		\cmidrule(lr){1-4}\cmidrule(lr){5-5}
		36.5 & 36.5                    & 38.4 & 37.5  & \textbf{49.5} \\ \bottomrule
	\end{tabular}
	\caption{\label{tab:bleu-eurollm-compare-adaptation-wiki}\BLEU score of various \EuroLLM-9B based systems on the Wikipedia test set ($0$-shot and $3$-shot). Target examples are retrieved with \LaBSE{} with a custom threshold of~$0.6$.}
\end{table}

We observe the superiority of the fine-tuning strategy compared to the various prompt-based approaches in Table~\ref{tab:bleu-eurollm-compare-adaptation-wiki}. As some of these differences might be explained by the implicit language and domain adaptation that takes place during model fine-tuning, we additionally fine-tune the base model with a \emph{standard} $k$-shot prompt, including the source and target side of each examples. This setting directly compares with the \emph{instruct} model, the only difference being the adaptation that is performed during fine-tuning; it also allows us to assess the performance loss that happens when using only target side examples. Results are in Table~\ref{tab:bleu-eurollm-compare-adaptation-ft-instruct} for a mixture of $5$ domains.

\begin{table}[!h]
	\centering
	\begin{tabular}{lllr}
		\toprule
		\EuroLLM model        & source    & k & \BLEU \\
		\midrule
		\multirow{2}{*}{Instruct}     & -         & 0 & 42.9  \\
		     & \ding{51} & 3 & 54.0  \\
		\midrule
		\multirow{2}{*}{\ft\texttt{base} (src+tgt)} & -         & 0 & 46.7  \\
		 & \ding{51} & 3 & 58.5  \\
		\midrule
		\multirow{2}{*}{\ft\texttt{base} (tgt)}     & -         & 0 & 47.3  \\
		    & \ding{56} & 3 & 57.5  \\
		\bottomrule
	\end{tabular}
	\caption{\label{tab:bleu-eurollm-compare-adaptation-ft-instruct}Average \BLEU score of various \EuroLLM-9B system fine-tuning schemes for ECB, EMEA, Europarl, JRC-Acquis and Wikipedia test sets (k-shot). Parallel examples are retrieved with \fuzzysrc{} using a custom threshold of~$0$.}
\end{table}

In Table~\ref{tab:bleu-eurollm-compare-adaptation-ft-instruct}, we observe that in fact the domain/language adaptation of \EuroLLM accounts for about +4 \BLEU points (\ft{base} (src+tgt) vs. instruct). For the sole Wikipedia domain, this adapation yields about +6 \BLEU points improvement. This confirms that method (b) is in fact clearly superior to simple prompting strategies. Therefore, this is the method used for all the results reported in Section~\ref{sec:results}. 
Moreover, we observe that the absence of the source side of $k$-shot examples is slightly detrimental to the model performance ($\approx$ -1 \BLEU point between the two fine-tuned versions, both for the $0$-shot and $3$-shot settings), suggesting that the target side of the retrieved examples contains most of the information needed to guide the model towards the desired translation.

% \todo{J'ai réécrit ce que je comprenais. Au vu de ces chiffres, la méthode la plus prometteuse est de faire un fine-tuning du modèle instruct (plutot que base). Quand tu parles de la méthode (b). Il serait peut-être plus naturel, vu qu'on a tous ces résultats, de simplifier un peu le discours: dans les scénarios avec prompting, il est attendu de prendre le modèle instruct, qui est fait pour cela; pour les scénarios avec FT, on peut continuer à utiliser le modèle base pour mesurer l'effet de pure adaptation, mais pour les expériences réelles il faut garder le modèle instruct - c'est ce qui semble le plus naturel.}
% \maxSmallTodo{Réponse au commentaire (commenté) : en fait j'ai fait une erreur dans le tableau en mettant "Base" au lieu de "Instruct". Dans tous les cas, le fine-tuning est fait sur le modèle base, et non pas sur le modèle instruct. J'avais eu des soucis pratiques avec le modèle instruct, ce qui fait que j'ai uniquement fine-tuné le modèle base. Mais Josep, je crois, a fait des expés où on voyait aucune différence de résultats entre les deux.}

\section{Computational Cost of Retrieval}

The computational cost can be dived into two categories: a fixed cost (independant from the number of sentences to translate) and a variable cost (proportional to this number). The retrieval process for training falls into the former, as well as the encoding of the whole set of monolingual target segments (for CLIR techniques) and the indexation of the TM (for fuzzy matching techniques). The variable costs lie in the fuzzy match search with filtering and the computation of \Lev, and in the CLIR search with the encoding of the source sentences, then the $k$NN search. In our experiments, we removed the filter (BM25) so that we obtain the best possible fuzzy matches (\fuzzysrc, \fuzzybt and \fuzzygold), at the cost of a $\sim$100 times higher latency. However, when applying a BM25 filter to preselect 100 segments, we observe that both fuzzy matching and CLIR techniques can retrieve their closest match in about $\sim$1ms. We obtained this result un corpus ECB (150K sentences) with optimmal conditions:
\begin{itemize}
	\item the fuzzy matching is performed on a 8 cores CPU, thus parallelizing the search on 8 simultaneous source sentences;
	\item as for CLIR, source sentences are encoded in batches of size 50, and the $k$NN search is handled by the FAISS library. We used a V100-32GB GPU.
\end{itemize}

% For training, retrieval is performed offline, once for each source sentence, before the training process begins.
% Before inference, the TM is indexed for fuzzy matching and the target monolingual dataset is encoded once for each CLIR embbeder.
% During inference, running fuzzy matching retrievers only requires to tokenize the source sentence $\src$ and compute BM25 w.r.t. all examples in the TM, then \Lev{} on a small subset of the 100 most relevant segments.
% Regarding the cross-lingual dense retrievers, they must first encode $\src$, then perform retrieval with a $k$NN search.
% We compared both techniques on ECB (150K) with optimal configurations (1 CPU with 8 cores for fuzzy matching;\footnote{The search for the closest examples is multithreaded on the 8 cores.} a V100-32GB GPU with batching\footnote{The encoding step is performed in parallel for batches of $50$ segments. Parallelisation of search is handled by the Faiss library.} for CLIR). Each method can retrieve the 1st best match in $\sim{1}$ms on average.

\section{Retrieval and Translation Examples \label{sec:examples}}

\begin{table*}[t]
	\resizebox{\textwidth}{!}{
		\begin{tabular}{lllp{0.7\textwidth}}
			\toprule

			                           & retriever               & \Lev{}                &                                                                                                                                                                                                        \\
			%  \cmidrule(lr){2-2}\cmidrule(lr){3-3}
			\hline
			\multirow{2}{*}{reference} &                         &                       & Le mannequin apparaît dans plusieurs séries  télévisées au début des années 2000. (\textsl{\footnotesize The model also appears in several TV shows in the early 2000s.})                              \\

			\cmidrule(lr){2-2}\cmidrule(lr){3-3}
			train                      & \LaBSE                  & 0.14                  & Cette série fut diffusée sur TF1 pendant l'été \textbf{2000.}  (\textsl{\footnotesize This show was broadcast on TF1 during the summer of 2000.})                                                      \\

			train                      & \ft\LaBSE\MAE           & 0.23                  & Elle est aussi apparue dans des \textbf{séries télévisées.}  (\textsl{\footnotesize She also played in some TV shows.})                                                                                \\

			\cmidrule(lr){2-2}\cmidrule(lr){3-3}
			mono                       & \LaBSE                  & 0.46                  & Elle se fait connaître par sa série \textbf{au début des années 2000.} (\textsl{\footnotesize She made a name for herself thanks to her show in the early 2000s.})                                     \\
			mono                       & \ft\LaBSE\MAE           & 0.62                  & Elle débute dans des \textbf{séries télévisées au début des années} 1990\textbf{.} (\textsl{\footnotesize She made her debut in some TV Shows in the early 1990s.})                                    \\
			\hline
			reference                  &                         &                       & Shasta a connu une histoire explosive et éruptive. (\textsl{\footnotesize Shasta has had an explosive and eruptive history.})                                                                          \\                               \\
			\cmidrule(lr){2-2}\cmidrule(lr){3-3}
			\multirow{2}{*}{train}     & \multirow{2}{*}{\LaBSE} & \multirow{2}{*}{0.06} & L'\textbf{histoire} de la région du Huaynaputina est marquée par un magma riche en silice\textbf{.}   (\textsl{\footnotesize The history of the Huaynaputina region is marked by silica-rich magma. }) \\
			train                      & \ft\LaBSE\MAE           & 0.44                  & Elle \textbf{a une histoire} très singulière\textbf{.}  (\textsl{\footnotesize She has a very peculiar history}.)                                                                                      \\
			\cmidrule(lr){2-2}\cmidrule(lr){3-3}
			mono                       & \LaBSE                  & 0.07                  & Il s'agit d'un des cônes volcaniques du mont \textbf{Shasta.}   (\textsl{\footnotesize This is one one the vulcanic cones of Mount Shasta.})                                                           \\
			mono                       & \ft\LaBSE\MAE           & 0.56                  & Elle \textbf{a connu une histoire} riche en rebondissements. (\textsl{\footnotesize Its history is full of twists and turns.})                                                                         \\
			\bottomrule
		\end{tabular}
	}
	\caption{\label{tab:illustration-example}Two illustrations of retrieved examples for the Wikipedia test set w.r.t. the chosen cross-lingual retriever, and the retrieval pool (train set vs. monolingual corpus). Exact matches are marked in \textbf{bold}. Indicative translations of each French segment into English are also provided.}
\end{table*}

An illustration of the retrieved examples is provided is table~\ref{tab:illustration-example}. It highlights the enhanced capability of \ft\LaBSE\MAE to retrieve lexically closer examples, rather than semantically close ones. This is particularly clear in the second example, which mentions Shasta, a volcano. While \LaBSE{} retrieves topic-related segments, \ft\LaBSE\MAE tends to select off-topic sentences that nonetheless contain a higher number of shared lexical items.

\section{Back-Translation \label{sec:backtrans}}

Back-translation (BT) has long been identified as a very effective technique to handle monolingual data in Statistical Machine-Translation \citep{bojar-tamchyna-2011-improving}, then in Neural Machine Translation \citep{sennrich-etal-2016-improving,currey-etal-2017-copied,edunov-etal-2018-understanding,burlot-yvon-2018-using,caswell-etal-2019-tagged}. Assuming the task is to translate from L1 into L2, and that monolingual texts in L2 are available, the BT approach translates these ``backward'' into L1, to obtain an artificial parallel corpus, pairing automatically generated source sentences with actual human-written target sentences. In NMT, it is custom to jointly use ``natural'' and ``artificial'' subsets of data to train and/or adapt translation models. This is illustrated in the first two columns (a) and (b) in Table~\ref{tab:btfury}, using no-retrieval ($k=0$), and corresponding respectively to the default (only parallel data) and BT-augmented settings.\footnote{We omit the case where only artificial data is used, as in fully unsupervised machine translation \citep{lample-etal-2018-phrase}.}

\begin{table}[h!]
	\centering
	\begin{tabularx}{1.0\linewidth}{r*{8}{c}}
		\toprule
		MT setting        & (a)       & (b)       & (c)       & (d)       & (e)                        & (f)       &  & \\
		\cmidrule(lr){1-1}\cmidrule(lr){2-7}
		\multicolumn{8}{l}{\bf Parallel data}                                                                           \\
		train ($k=0$)     & \ding{51} & \ding{51} & -         & -         & \textcolor{red}{-}         & -         &  & \\
		train ($k\geq 0$) & -         & -         & \ding{51} & \ding{51} & \textcolor{red}{\ding{51}} & \ding{51} &  & \\
		test  ($k\geq 0$) & -         & -         & \ding{51} & \ding{51} & \textcolor{red}{\ding{51}} & \ding{51} &  & \\
		%     \multicolumn{9}{l}{\bf Monolingual data, Back-Translated} \\
		\cmidrule(lr){1-1}\cmidrule(lr){2-7}
		\multicolumn{9}{l}{\bf BT-ed data}                                                                              \\
		train ($k=0$)     & -         & \ding{51} & -         & -         & \textcolor{red}{-}         & -         &  & \\
		train ($k\geq 0$) & -         & -         & -         & \ding{51} & \textcolor{red}{-}         & \ding{51} &  & \\
		test  ($k\geq 0$) & -         & -         & -         & -         & \textcolor{red}{\ding{51}} & \ding{51} &  & \\
		\bottomrule
	\end{tabularx}
	\caption{Using back-translation in NMT and RANMT}
	\label{tab:btfury}
\end{table}

Column (c) in Table~\ref{tab:btfury} illustrates standard RANMT, where $k \geq 0$ parallel examples are used during training and at inference; columns (d), (e) and (f) correspond to three possible ways to integrate artificial data: (d) only during training, and retrieve selectively from the high-quality parallel data; (e) only during inference, extending a generic RANMT with relevant data; (f) both during training and at inference as in \cite{tezcan-etal-2024-improving}. Our use of back-translation in the experiments of Section~\ref{sec:experiments} corresponds to column (e), a setting we deem representative of typical use of monolingual data in actual applications, and which crucially does not require retraining stage. Note that the same scenario is used in our CLIR approach, which, in addition, fully dispenses with the back-translation stage.

\section{Experiments in English-Ukrainian \label{sec:en2uk}}

To further emphasize the ability of our method to retrieve relevant examples even in the absence of direct lexical overlap between source and target segments, we experiment with the translation from English into Ukrainian, with the source using the Latin script, and the target the Cyrillic script.  We used the same dataset as \citet{tezcan-etal-2024-improving}, corresponding to legal texts. The parallel corpus is split into train, validation and test sets, each containing respectively 286,417, 2000 and 1898 sentences. This corpus is augmented with a monolingual in-domain dataset of 1,461,320 Ukrainian sentences.

Our experiments are conducted with the same \NFA architecture as the experiments in Section~\ref{sec:experiments}, as it produces translations with higher estimated-quality than the non-autoregressive system. This \NFA model has been trained on 5M en-uk parallel pairs from various domains, including the in-domain train set.

\begin{table}[h!]
	\resizebox{\linewidth}{!}{
		\begin{tabular}{llrr}
			\toprule
			Retrieval method & Corpus     & \BLEU    & \COMET \\
			\hline
			no example       & -          & 51.2     & 92.7   \\
			\fuzzysrc        & train      & 54.2     & 92.8   \\
			\LaBSE           & train      & 53.9     & 92.7   \\
			\ft\LaBSE\MSE    & train      & $^*$54.3 & 92.8   \\
			\LaBSE           & train+mono & 57.2     & 92.9   \\
			\ft\LaBSE\MSE    & train+mono & $^*$57.7 & 93.0   \\
			\bottomrule
		\end{tabular}
	}
	\caption{\label{tab:nfa-en-uk} \BLEU and \COMET scores obtained by the \NFA model according to the retrieval method. Significance w.r.t. to the \LaBSE{} counterpart is indicated by $^*$.}
\end{table}

We fine-tune \LaBSE{} using the proposed MSE lexical loss, with examples retrieved via \fuzzysrc from the training set. We then evaluate and compare the performance of \fuzzysrc, \LaBSE, and \ft\LaBSE\MSE by computing \BLEU and \COMET scores of translations produced by the \NFA model. For \LaBSE{} and \ft\LaBSE\MSE, we consider two retrieval settings — \textit{train} and \textit{train+mono} — which correspond to the datasets from which target-side examples are retrieved for the CLIR approach. Results are in Table~\ref{tab:nfa-en-uk}, scores are computed respectively by SacreBLEU and \COMET-22, significance tests also use the SacreBLEU implementations with paired bootstrap resampling (n=1000, p=0.05).

We can make the following observations: (a) the baseline \NFA model is about 3 BLEU points better than baseline encoder-decoder model; (b) CLIR here again matches the default RAMT set-up, where retrieval only exploits the parallel training data; (c) augmenting the retrieval pool with monolingual data again induces a significant performance boost (+3.4 \BLEU), when retrieval relies on the fine-tuned version of \LaBSE. These observations are consistent with what is reported in the main text.

\section{\COMET Scores \label{sec:comet}}

For the full picture, we also report the \COMET scores of each domain for all our experiments in Tables~\ref{tab:comet-per-domain-mlevt-clir}, \ref{tab:comet-per-domain-nfa-clir} and \ref{tab:comet-per-domain-eurollm-clir}.

\begin{table*}[ht]
	\centering
	\small
	\resizebox{\textwidth}{!}{
		\begin{tabular}{lrrrrrrrrrrrrrrrr}
			\toprule
			% & \multicolumn{11}{c||}{English-French} & \multicolumn{5}{c|}{German-English}                                                                                                                                                                                                                                                                                                                                                                                                                                                    \\
			                    & \multicolumn{11}{c}{English-French} & \multicolumn{5}{c}{German-English}                                                                                                                                                                                                                                                                                                                                                                                                                                                     \\[-1pt]
			\cmidrule(lr){2-12}\cmidrule(lr){13-17}
			~                   & \makebox[\widthof{xxx}]{ECB}        & \makebox[\widthof{xxx}]{EME}       & \makebox[\widthof{xxx}]{Epp} & \makebox[\widthof{xxx}]{GNO} & \makebox[\widthof{xxx}]{JRC} & \makebox[\widthof{xxx}]{KDE} & \makebox[\widthof{xxx}]{News} & \makebox[\widthof{xxx}]{PHP} & \makebox[\widthof{xxx}]{TED} & \makebox[\widthof{xxx}]{Ubu} & \makebox[\widthof{xxx}]{Wiki} & \makebox[\widthof{xxx}]{KDE} & \makebox[\widthof{xxx}]{Kor} & \makebox[\widthof{xxx}]{JRC} & \makebox[\widthof{xxx}]{EME} & \makebox[\widthof{xxx}]{Sub} \\[-2pt] \cmidrule(lr){1-12} \cmidrule(lr){13-17} \cmidrule(lr){1-12} \cmidrule(lr){13-17}
			\tn{\fuzzygold}     & 70.5 & 85.0 & 45.8 & 73.2 & 71.9 & 58.9 & 24.3 & 35.9 & 21.2 & 52.1 & 29.8 & 30.3 & -98.3  & 19.9  & 29.7 & -24.8 \\[-2pt]
			\tn{\fuzzysrc}      & \bf 69.0 & \bf 81.5 & 44.7 & \bf 67.6 & \bf 69.8 & 47.9 & \bf 24.7 & \bf 33.5 & 19.5 & 49.2 & \bf 25.0 & 24.1 & -85.8  & 16.6  & 21.8 & -28.2 \\[-2pt]
			\tn{\fuzzybt}       & 61.6 & 75.0 & 45.6 & 55.5 & 65.8 & 37.3 & 24.1 & 30.1 & 17.7 & 45.4 & 24.2 & 11.4 & -110.6 & -11.0 & 21.3 & \bf -25.4 \\[-1pt] \cmidrule(lr){1-12} \cmidrule(lr){13-17}
			\tn{\LASER}         & 65.8 & 76.3 & 45.1 & 58.1 & 66.0 & 34.7 & 23.2 & 29.2 & \bf 19.9 & 45.0 & 24.5 & 18.5 & -73.0  & 19.7  & \bf 27.5 & -31.0 \\[-2pt]
			\tn{\LaBSE}         & 65.5 & 77.2 & 42.3 & 62.9 & 67.9 & 42.8 & 23.3 & 29.7 & 19.0 & 47.1 & 23.3 & \bf 21.2 & -86.4  & \bf 21.4  & 26.3 & -30.1 \\[-2pt] \cmidrule(lr){1-12} \cmidrule(lr){13-17}
			\tn{\ft\LaBSE\MSE } & 68.0 & \bf 81.1 & 44.4 & \bf 66.5 & \bf 69.8 & \bf 48.0 & \bf 24.4 & 31.9 & 18.5 & 49.6 & 23.8 & 18.5 & -76.3  & 8.7   & 15.6 & -29.1 \\[-2pt]
			\tn{\ft\LaBSE\MAE } & \bf 69.0 & 80.7 & 44.2 & 65.5 & 68.4 & 47.0 & 23.8 & \bf 32.7 & 18.9 & \bf 49.8 & \bf 24.9 & 20.2 & \bf -70.6  & 14.1  & 19.1 & -28.6 \\[-2pt]
			\tn{\ft\LaBSE\Rank} & 67.3 & 79.6 & \bf 44.8 & 63.4 & 68.8 & 46.5 & 23.8 & 32.2 & 19.7 & 47.9 & 24.5 & 20.7 & -84.9  & 14.0  & 17.6 & -28.5 \\[-1pt]
			\bottomrule
		\end{tabular}
	}
	\caption{\label{tab:comet-per-domain-mlevt-clir} Per domain \COMET scores for \mlevt{3}.}
\end{table*}

\begin{table*}[t]
	\centering
	\small
	\resizebox{\textwidth}{!}{
		\begin{tabular}{lrrrrrrrrrrrrrrrr}
			\toprule
			% & \multicolumn{11}{c||}{English-French} & \multicolumn{5}{c|}{German-English}                                                                                                                                                                                                                                                                                                                                                                                                                                                    \\
			                    & \multicolumn{11}{c}{English-French} & \multicolumn{5}{c}{German-English}                                                                                                                                                                                                                                                                                                                                                                                                                                                     \\[-1pt]
			\cmidrule(lr){2-12}\cmidrule(lr){13-17}
			~                   & \makebox[\widthof{xxx}]{ECB}        & \makebox[\widthof{xxx}]{EME}       & \makebox[\widthof{xxx}]{Epp} & \makebox[\widthof{xxx}]{GNO} & \makebox[\widthof{xxx}]{JRC} & \makebox[\widthof{xxx}]{KDE} & \makebox[\widthof{xxx}]{News} & \makebox[\widthof{xxx}]{PHP} & \makebox[\widthof{xxx}]{TED} & \makebox[\widthof{xxx}]{Ubu} & \makebox[\widthof{xxx}]{Wiki} & \makebox[\widthof{xxx}]{KDE} & \makebox[\widthof{xxx}]{Kor} & \makebox[\widthof{xxx}]{JRC} & \makebox[\widthof{xxx}]{EME} & \makebox[\widthof{xxx}]{Sub} \\[-2pt]
			\cmidrule(lr){2-12}\cmidrule(lr){13-17}
			\sm{no example}     & 58.9 & 55.9 & 39.9 & 48.8 & 62.3 & 42.8 & 50.1 & 45.6 & 43.9 & 43.5 & 36.4 & 82.2 & 68.9 & 83.2 & 84.7 & 80.2 \\[-2pt] \cmidrule(lr){1-12} \cmidrule(lr){13-17}
			\tn{\fuzzygold}     & 90.5 & 90.9 & 87.9 & 89.8 & 90.7 & 85.8 & 88.2 & 84.6 & 85.4 & 86.6 & 85.2 & 84.5 & 71.4 & 88.2 & 86.3 & 80.6 \\[-2pt]
			\tn{\fuzzysrc}      & \bf 90.4 & \bf 90.5 & \bf 87.8 & \bf 89.3 & \bf 90.5 & 85.2 & \bf 88.2 & \bf 84.5 & \bf 85.4 & 86.3 & \bf 85.0 & \bf 84.1 & 70.0 & \bf 87.9 & \bf 85.9 & \bf 80.4 \\[-2pt]
			\tn{\fuzzybt}       & 90.0 & 90.1 & 87.7 & 88.6 & 89.9 & 84.4 & \bf 88.2 & 84.3 & 85.2 & 86.0 & \bf 84.9 & 83.5 & 70.1 & 87.5 & 85.2 & 80.3 \\[-1pt] \cmidrule(lr){1-12} \cmidrule(lr){13-17}
			\tn{\LASER}         & 90.0 & 90.1 & \bf 87.8 & 88.8 & 90.0 & 84.4 & \bf 88.2 & 84.2 & \bf 85.4 & 86.4 & 84.7 & 83.5 & 70.0 & 87.4 & 85.3 & \bf 80.4 \\[-2pt]
			\tn{\LaBSE}         & 90.0 & 90.3 & 87.5 & 89.0 & 89.8 & 84.7 & \bf 88.2 & 84.1 & 85.3 & 86.4 & \bf 84.9 & \bf 83.8 & 70.3 & 87.2 & 85.5 & 80.3 \\[-2pt] \cmidrule(lr){1-12} \cmidrule(lr){13-17}
			\tn{\ft\LaBSE\MSE}  & \bf 90.3 & \bf 90.5 & 87.7 & \bf 89.2 & 90.4 & 85.1 & \bf 88.2 & \bf 84.5 & 85.3 & \bf 86.5 & 84.8 & \bf 83.8 & 70.3 & \bf 87.8 & 85.6 & \bf 80.4 \\[-2pt]
			\tn{\ft\LaBSE\MAE}  & \bf 90.3 & \bf 90.5 & 87.7 & \bf 89.2 & 90.4 & 85.1 & \bf 88.2 & \bf 84.5 & \bf 85.4 & 86.4 & 84.8 & 83.7 & 70.2 & 87.7 & \bf 85.7 & 80.3 \\[-2pt]
			\tn{\ft\LaBSE\Rank} & 90.2 & 90.4 & 87.7 & \bf 89.2 & \bf 90.5 & \bf 85.3 & \bf 88.2 & \bf 84.5 & 85.3 & \bf 86.5 & \bf 84.9 & \bf 83.8 & \bf 70.5 & \bf 87.8 & \bf 85.7 & \bf 80.4 \\[-1pt]
			\bottomrule
		\end{tabular}
	}
	\caption{\label{tab:comet-per-domain-nfa-clir} Per domain \COMET scores for \NFA.}
\end{table*}

\begin{table*}[ht]
	\centering
	\small
	% \begin{tabular}{|l||rrrrrrrrrrr||rrrrr|}
	\resizebox{\textwidth}{!}{
		\begin{tabular}{lrrrrrrrrrrrrrrrr}
			\toprule
			% & \multicolumn{11}{c||}{English-French} & \multicolumn{5}{c|}{German-English}                                                                                                                                                                                                                                                                                                                                                                                                                                                    \\
			                    & \multicolumn{11}{c}{English-French} & \multicolumn{5}{c}{German-English}                                                                                                                                                                                                                                                                                                                                                                                                                                                     \\[-1pt]
			\cmidrule(lr){2-12}\cmidrule(lr){13-17}
			~                   & \makebox[\widthof{xxx}]{ECB}        & \makebox[\widthof{xxx}]{EME}       & \makebox[\widthof{xxx}]{Epp} & \makebox[\widthof{xxx}]{GNO} & \makebox[\widthof{xxx}]{JRC} & \makebox[\widthof{xxx}]{KDE} & \makebox[\widthof{xxx}]{News} & \makebox[\widthof{xxx}]{PHP} & \makebox[\widthof{xxx}]{TED} & \makebox[\widthof{xxx}]{Ubu} & \makebox[\widthof{xxx}]{Wiki} & \makebox[\widthof{xxx}]{KDE} & \makebox[\widthof{xxx}]{Kor} & \makebox[\widthof{xxx}]{JRC} & \makebox[\widthof{xxx}]{EME} & \makebox[\widthof{xxx}]{Sub} \\[-2pt] \cmidrule(lr){1-12} \cmidrule(lr){13-17} \cmidrule(lr){1-12} \cmidrule(lr){13-17}
			\sm{no example}     & 88.2 & 88.5 & 87.7 & 85.3 & 89.1 & 82.8 & 86.2 & 84.3 & 85.0 & 85.8 & 86.3 & 85.9 & 74.7 & 86.9 & 85.1 & 80.5 \\[-2pt]
			\tn{\fuzzygold}     & 90.2 & 90.9 & 87.7 & 90.1 & 90.3 & 87.5 & 86.3 & 85.2 & 85.4 & 89.0 & 87.0 & 88.0 & 75.5 & 88.4 & 87.0 & 80.8 \\[-2pt]
			\tn{\fuzzysrc}      & \bf 90.0 & 90.5 & \bf 87.8 & \bf 89.3 & \bf 90.1 & 86.2 & 86.2 & \bf 85.1 & \bf 85.3 & 88.4 & \bf 86.8 & 86.5 & 74.9 & 87.2 & 85.5 & \bf 80.7 \\[-2pt]
			\tn{\fuzzybt}       & 88.6 & 89.0 & \bf 87.7 & 87.2 & 89.2 & 84.0 & 86.2 & 84.7 & \bf 85.3 & 87.6 & 86.6 & 86.0 & 74.9 & 87.2 & 85.6 & \bf 80.6 \\[-1pt] \cmidrule(lr){1-12} \cmidrule(lr){13-17}
			\tn{\LASER}         & 89.8 & 90.0 & \bf 87.7 & 88.1 & 90.0 & 84.6 & \bf 86.3 & 84.9 & 85.2 & 88.3 & 86.6 & 86.3 & 75.0 & \bf 87.9 & \bf 85.9 & \bf 80.6 \\[-2pt]
			\tn{\LaBSE}         & \bf 90.0 & 90.2 & \bf 87.7 & 88.6 & \bf 90.1 & 85.8 & \bf 86.3 & 84.9 & 85.2 & 88.2 & 86.6 & 86.5 & 75.0 & \bf 87.9 & 86.1 & 80.5 \\[-2pt] \cmidrule(lr){1-12} \cmidrule(lr){13-17}
			\tn{\ft\LaBSE\MSE } & \bf 90.0 & 90.5 & \bf 87.7 & 88.9 & \bf 90.1 & \bf 86.4 & \bf 86.3 & \bf 85.1 & \bf 85.3 & 88.5 & \bf 86.8 & 86.5 & \bf 75.1 & 87.8 & 85.8 & \bf 80.6 \\[-2pt]
			\tn{\ft\LaBSE\MAE } & 89.9 & \bf 90.6 & \bf 87.7 & 89.1 & \bf 90.1 & \bf 86.4 & 86.2 & 85.0 & 85.2 & 88.4 & 86.7 & \bf 86.7 & \bf 75.1 & \bf 87.9 & 86.0 & \bf 80.5 \\[-2pt]
			\tn{\ft\LaBSE\Rank} & 89.8 & 90.3 & \bf 87.7 & \bf 89.2 & 90.0 & 86.2 & \bf 86.3 & 84.9 & \bf 85.3 & \bf 88.6 & 86.6 & 86.6 & 75.0 & 87.8 & 86.0 & 80.5 \\[-1pt]
			\bottomrule
		\end{tabular}
	}
	\caption{\label{tab:comet-per-domain-eurollm-clir} Per domain \COMET scores for \EuroLLM.}
\end{table*}

\end{document}